\documentclass[letterpaper,journal]{IEEEtran}
\usepackage{amsmath,amsfonts}
\usepackage{algorithmic}
\usepackage{algorithm}
\usepackage{array}
\usepackage[caption=false,font=normalsize,labelfont=sf,textfont=sf]{subfig}
\usepackage{textcomp}
\usepackage{stfloats}
\usepackage{url}
\usepackage{verbatim}
\usepackage{graphicx}
\usepackage{booktabs}       % professional-quality tables
\usepackage{tabularx}
\usepackage{colortbl}
\usepackage{float}
\usepackage{fancyhdr}       % header
\usepackage{ragged2e} % Better hyphenation in narrow columns
\usepackage[backend=biber, style=numeric-verb, sorting=none]{biblatex}
\DeclareBibliographyAlias{standard}{manual}

\usepackage[hidelinks]{hyperref} % hyperlinks (hides ugly red boxes)
\usepackage[nohyperlinks,printonlyused]{acronym}
\newcolumntype{L}{>{\RaggedRight\arraybackslash}X}
\graphicspath{{./figures}}
\begin{document}

\title{CARLAverse: A Highly Modular, Distributed, and Multimodal Framework for Human-in-the-Loop Simulation}

\author{Patrick Rebling, Philipp Nenninger, Reiner Kriesten \\
	Institute of Energy Efficient Mobility \\
	Karlsruhe University of Applied Sciences \\
	\texttt{\{patrick.rebling, philipp.nenninger, reiner.kriesten\}@h-ka.de}
        % <-this % stops a space
}

% The paper headers
\markboth{}%
{Rebling \MakeLowercase{\textit{et al.}}: CARLAverse}

\maketitle

\begin{abstract}
	The development of autonomous driving demands comprehensive testing in mixed-traffic scenarios involving vulnerable road users (VRUs), where purely artificial agents often fail to capture authentic human social negotiations. While human-in-the-loop (HITL) simulators enable safe investigation of these interactions, existing multi-agent platforms struggle with the network latency and synchronization constraints required for high-fidelity haptic feedback. To resolve this, we present CARLAverse, an open-source, multimodal simulation ecosystem. Extending modular hardware abstraction, CARLAverse integrates driving (DrivoCARLA), cycling (CycloCARLA), and pedestrian (WalkoCARLA) simulators into a shared virtual environment. Its core methodological contribution is a distributed physics architecture: latency-critical ego dynamics and high-frequency force feedback are computed locally on client nodes, while a central CARLA server orchestrates non-player character (NPC) physics and global traffic. By decoupling haptic control loops from network bottlenecks, CARLAverse enables scalable, cross-institutional HITL experiments without compromising physical immersion. Code and documentation: \url{https://git.ieem-ka.de/simulator-environments/carlaverse}.
\end{abstract}


\acresetall

\section{Introduction}
\label{sec:intro}

Over the past decade, ambitious timelines and predictions have driven the development of autonomous driving \cite{muskInterviewElonMusk2016,bimbrawAutonomousCarsPresent2015}. Despite considerable technological progress, achieving fully autonomous vehicles capable of reliably navigating all potential driving conditions remains a persistent challenge \cite{padmajaExplorationIssuesChallenges2023}. One major challenge is the variety of mixed traffic scenarios in which automated vehicles must coexist with human-driven cars and \acp{VRU}, such as cyclists and pedestrians. Consequently, observing, predicting, and accurately modeling human behavior within these complex, dynamic environments has become a pivotal focus of autonomous driving research. Addressing this issue is essential not only for operational safety, but also for gaining broader public acceptance \cite{kellyWhatFactorsContribute2023} and trust \cite{gillespieTrustArtificialIntelligence2023} in \ac{AI}-driven transportation systems \cite{molnarUnderstandingTrustAcceptance2018}. In this context, \ac{HITL} simulation is a crucial methodology that enables the safe and repeatable investigation of human behavior in complex traffic interactions without exposing participants to physical risk.

However, effectively deploying \ac{HITL} driving simulators introduces significant technical challenges. Ecological validity relies heavily on the system's ability to achieve a high degree of realism. Visual, auditory, and tactile feedback must be closely synchronized and authentic to generate natural responses from drivers \cite{sawadaEffectsSynchronisedEngine2020, kemenyEvaluatingPerceptionDriving2003}. Furthermore, minimizing latency between user input and simulator response is essential. Delays can induce simulator sickness, disrupt the human sensorimotor loop, and compromise the integrity of the collected behavioral data. Finally, integrating heterogeneous hardware and software subsystems is challenging due to the varying communication protocols and data standards of input and output devices.

To mitigate the complexity of constructing individualized simulator setups, our previous work introduced a modular hardware abstraction framework based on the \ac{CARLA} simulation environment \cite{reblingHighlyModularImmersiveHumanintheLoop2025}. While this architecture successfully reduces the overhead associated with deploying standalone, immersive simulators, a comprehensive study of mixed traffic requires observing concurrent interactions among multiple human participants. Existing solutions to this multi-agent problem often have limitations: commercial connected simulators are usually too expensive \cite{silvaRealistic3DSimulators2024, rodsethNovelLowcostSolution2017}, and current open-source alternatives usually lack the necessary immersion and streamlined hardware integration for a cohesive multi-actor ecosystem \cite{ihemedu-steinkeDevelopmentEvaluationVirtual2015, azimiSimulationAllStepbyStep2025}. Therefore, there is a persistent need for a comprehensive simulation framework that can seamlessly integrate various road users into a shared, responsive virtual environment \cite{azimiSimulationAllStepbyStep2025, creweSLAVSimFrameworkSelfLearning2023}.

To address this gap, this paper introduces CARLAverse, an open, multimodal, and connected simulation ecosystem. The primary contribution of this work is the overarching system architecture, software framework, and cross-domain integration methodology that enables heterogeneous \ac{HITL} nodes to interact within a shared virtual environment. While the paper includes technical descriptions of individual subsystems (including bicycle dynamics, spatial audio, force-feedback haptics, motion cueing, pedestrian \ac{VR}, and vision-based pose estimation), these sections serve to document their integration interfaces, data pipelines, and operational system roles within the distributed ecosystem. CARLAverse integrates various \ac{HITL} simulators within a unified \ac{CARLA}-based infrastructure. Expanding upon our previously established modular architecture, CARLAverse currently supports three interaction domains: vehicle (DrivoCARLA), bicycle (CycloCARLA), and pedestrian (WalkoCARLA) simulators. A core methodological contribution of this ecosystem is its distributed physics architecture. To support high-fidelity force feedback and low-latency interaction for human participants in complex urban simulations, physics computation is structurally decoupled. Each participant's node-specific ego physics is calculated locally on dedicated local nodes, while the central \ac{CARLA} server manages \acp{NPC} physics and centrally coordinated global traffic simulation. This distributed paradigm addresses the trade-off between localized haptic realism and centralized environmental synchronization, and is intended to support scalable, connected, multi-agent experiments. This paper documents the architecture, implementation, and integration principles of CARLAverse. Detailed synchronization mechanisms and quantitative validation of individual physical and perceptual subsystems are addressed in dedicated companion publications and ongoing work.

The remainder of this paper is organized as follows: Section \ref{sec:soa} reviews the state of the art and related work in connected simulation frameworks. Section \ref{sec:carlaverse} details the distributed software architecture and networking paradigm of the CARLAverse. Section \ref{sec:crosscut} outlines cross-cutting technological integrations, such as the CARLAcoustics audio engine. Section \ref{sec:simdetail} presents the individual multimodal simulation nodes, specifically DrivoCARLA, WalkoCARLA, and CycloCARLA. Finally, Section \ref{sec:con} summarizes the findings and highlights directions for future research.

\section{Related Work and State of the Art}
\label{sec:soa}

This section reviews existing approaches in hardware abstraction, the current state of mixed-traffic simulation, and the technical challenges associated with distributed physics and latency management in multi-agent environments to contextualize the framework of the CARLAverse ecosystem.

\subsection{Modular Software and Hardware Abstraction}

The modularization of software for the purpose of achieving hardware abstraction is an established paradigm designed to ensure flexibility and reusability. The implementation of \acp{HAL} \cite{massaHardwareAbstractionLayer2003} facilitates the development of reconfigurable software architectures. Virtual machines are a prominent \ac{HAL} technology that simulate operating systems across diverse host environments \cite{liSurveyVirtualMachine2010}. The \ac{ROS} \cite{quigleyROSOpensourceRobot2009} is extensively used for hardware abstraction, enabling modular development where hardware-specific components can be seamlessly interchanged. In the automotive sector, the \ac{AUTOSAR} standard \cite{vogetAUTOSARAutomotiveTool2010} provides a formalized methodology for hardware abstraction. Furthermore, recent advancements have introduced \acp{HAL} for embedded systems that use time-triggered hardware access to ensure deterministic execution \cite{simmannDesignAlternativeHardware2024}.

On the hardware level, modularization has traditionally been achieved through interchangeable mock-up modules \cite{defilippoModularArchitectureDriving2014}. For instance, Realtime Technologies offers RDS-Modular simulator mock-ups \cite{technologiesRDSModular2024}, which allow for customized assembly from predefined components. However, commercial simulation solutions typically present a high financial barrier to entry and rely on proprietary interfaces, which can restrict users to specific vendor ecosystems for hardware integration \cite{silvaRealistic3DSimulators2024, rodsethNovelLowcostSolution2017}. Conversely, while modern open-source frameworks deliver high visual fidelity, they frequently lack native support for physical immersion (e.g., motion platforms or force-feedback systems) and necessitate extensive source code modifications to interface with customized hardware \cite{ihemedu-steinkeDevelopmentEvaluationVirtual2015}.

\subsection{Open-Source Simulators and Mixed-Traffic Integration}

Simulation environments, such as \ac{CARLA} in the open-source domain or the driving simulation software SILAB (WIVW GmbH) \cite{kruegerSILABTaskOriented2005} in the commercial sector, typically use a modular architecture focused on software interfaces for testing autonomous driving functions and \acp{ADAS}. For driving simulators, MATLAB/SIMULINK-based approaches have been employed to customize vehicle dynamics within \ac{HITL} setups \cite{cipelliDriverintheloopSimulationsParametric2008}. While modular software-based architectures exist for model-based testing of automated driving functions \cite{fischerModularScalableDriving2014}, they are often optimized for specific simulator configurations or lack open-source accessibility.

The academic community heavily relies on open-source platforms to establish reproducible testing environments, with the landscape largely divided by application scale. Microscopic traffic flow simulators like Eclipse \ac{SUMO} \cite{lopezMicroscopicTrafficSimulation2018} are highly efficient at simulating large-scale urban mobility and vehicle routing, yet they lack high-fidelity 3D rendering and complex ego-vehicle dynamics. Conversely, 3D simulators built upon game engines, such as \ac{LGSVL} simulator \cite{rongLGSVLSimulatorHigh2020}, AirSim \cite{shahAirSimHighFidelityVisual2017}, and \ac{CARLA}, have become standard for sensor simulation and \ac{AV} software testing. Frameworks like JOAN \cite{beckersJOANFrameworkHumanautomated2023} leverage \ac{CARLA} to provide excellent environments for single-ego \ac{HITL} setups. However, they are primarily engineered to test a single human driver or \ac{AV} algorithm against a computer-controlled environment.

Transitioning from single-ego to mixed-traffic scenarios highlights a primary limitation in current open-source simulators: the behavioral modeling of \acp{VRU}. In platforms like \ac{CARLA} or \ac{SUMO}, \acp{VRU} are typically represented as \acp{NPC} governed by rule-based algorithms, finite-state machines, or \acp{FM} \cite{rampfModellingAutonomousVehicle2023}. These artificial agents often exhibit a behavioral gap during complex social negotiations. Subtle dynamics such as eye contact, yielding behavior, body language, and hesitation \cite{sahaiCrossingStreetFront2022, quanteMeasuringDescribingCooperation2024} are nuanced human traits that current \ac{AI} systems struggle to replicate authentically.

To address this, recent research has used co-simulation, integrating \ac{CARLA} with \ac{SUMO} via \ac{TraCI} \cite{azfarTrafficCosimulationFramework2025, wegenerTraCIInterfaceCoupling2008}. While effective for orchestrating artificial agents, co-simulation does not fully address the synchronization constraints required for distributed \ac{HITL} testing, where multiple human participants require real-time haptic and vestibular feedback without experiencing simulator sickness \cite{ramlallDevelopmentNetworkedMultiParticipant2025}. Overcoming this gap requires multi-agent \ac{HITL} setups capable of placing multiple human participants in a shared virtual space to capture genuine interactions. Recent frameworks, such as \ac{CARMEn} \cite{machadoCARMEnCARlaBasedMultiAgent2026}, have introduced multi-agent environments focusing on \ac{VR} and spatial audio to facilitate driver-pedestrian coexistence. However, \ac{CARMEn} relies on a centralized architecture and standard middleware, which may introduce serialization latency that can affect the fidelity of mechanical feedback systems. A holistic approach must support heterogeneous nodes, such as pedestrian \ac{VR} simulators and physical bicycle trainers, while maintaining stringent physical synchronization.

To fully appreciate the structural requirements of the CARLAverse, it is essential to examine the broader landscape of multi-agent \ac{HITL} simulators outside the \ac{CARLA} ecosystem. As summarized in Table \ref{tab:azimi_comparison} (adapted from a recent review \cite{azimiSimulationAllStepbyStep2025}), several platforms have attempted to bridge the gap in multi-agent interactions. However, they share distinct limitations regarding scalability and hardware abstraction.

\begin{table*}[tbp]
	\caption{Summary of Non-CARLA Multi-Agent Simulation Studies and Frameworks (Adapted from \cite{azimiSimulationAllStepbyStep2025})}
	\label{tab:azimi_comparison}
	\centering
	\small
	\renewcommand{\arraystretch}{1.2}
	\setlength{\tabcolsep}{4pt}
	\begin{tabularx}{\textwidth}{l c L L L}
		\toprule
		\textbf{Study} & \textbf{Agents} & \textbf{Agent Types} & \textbf{Simulation Platforms} & \textbf{Architecture \& Hardware Dependency} \\
		\midrule
		Lehsing \& Feldstein (2018) \cite{lehsingUrbanInteractionGetting2018} & 2 & Driver, Pedestrian & Full cabin sim, CAVE & Centralized, proprietary hardware lock-in \\
		Perez et al. (2019) \cite{perezARPEDFrameworkAugmented2019} & 2 & Driver (Pod), Pedestrian & Desktop sim, \ac{VR} / AR & Centralized \\
		Lindner et al. (2022) \cite{lindnerCoupledDrivingSimulator2022} & 2 & \ac{AV}, Pedestrian & Desktop sim & Centralized \\
		Sabeti et al. (2024) \cite{sabetiMADIVEMultiAgentDistributed2024} & 3 & Driver, Pedestrian, Cyclist & Desktop sim, \ac{VR} & Localized distributed \ac{VR} \\
		Crosato et al. (2024) \cite{crosatoVirtualRealityFramework2024} & 2 & Driver, Pedestrian & Desktop sim, \ac{VR} & Centralized single-node physics \\
		Yang et al. (2024) \cite{yangUsingDistributedSimulations2024} & 2 & Driver, Pedestrian & Full cabin sim, CAVE & Distributed, hardware-specific \\
		Simulation for All (2025) \cite{azimiSimulationAllStepbyStep2025} & 4 & Driver, Ped., Cyclist, Transit & Motion Cockpit, Treadmill, Bike Trainer & Localized \ac{UDP}; hardware-specific \\
		\bottomrule
	\end{tabularx}
\end{table*}

First, the majority of these studies are limited to only two interacting agents (typically a driver and a pedestrian) operating within centralized, single-node physics environments. Second, while the recently proposed \textit{Simulation for All} framework \cite{azimiSimulationAllStepbyStep2025} significantly advances the field by integrating up to four agent types (including public transit), it remains fundamentally constrained by its network architecture and rigid hardware dependencies. The system mandates specific, physical infrastructure, such as proprietary omnidirectional treadmills and high-end \ac{VR} headsets, operating exclusively over a localized \ac{UDP} network. Consequently, it is not primarily designed for hardware-agnostic integration of custom simulator mock-ups. Furthermore, its localized architecture does not describe support for geographically distributed deployments across \acp{WAN} while maintaining rigorous haptic synchronization. This critical gap underscores the necessity for the distributed physics and modular abstraction paradigm introduced by the CARLAverse.

\subsection{Distributed Architectures and Latency Management}

Operating a multi-agent \ac{HITL} ecosystem introduces substantial computational and networking bottlenecks. For immersive simulators, mechanical, visual, and communication latencies must be strictly minimized to ensure valid behavioral data \cite{kemenyEvaluatingPerceptionDriving2003, saidiTransportDelayCharacterization2010}. Rendering visual environments, calculating multi-body physics, and computing force feedback necessitate extremely high frequencies (often \textgreater100\,Hz, up to 1000\,Hz for stable haptics) \cite{heStableHapticRendering2013}.

Historically, modular simulator frameworks have relied on middleware like \ac{ROS} or \ac{ROS}\,2 for node orchestration. However, latency profiling of distributed \ac{ROS} systems indicates structural bottlenecks for time-critical haptic applications \cite{kronauerLatencyAnalysisROS22021}. While higher transmission frequencies in \ac{ROS}\,2 can reduce median latency via active thread scheduling, the framework introduces considerable baseline computational overhead. Routing control signals through the \ac{ROS}\,2 stack and its underlying \ac{DDS} can add up to 50\,\% more latency compared to raw socket communication for small payloads. This compounding overhead scales linearly with node count, degrading the fidelity of mechanical systems such as direct-drive steering wheels or resistance motors. Traditional latency mitigation strategies, including predictive algorithms or robust control techniques, struggle with high-frequency dynamic feedback, such as self-aligning torque or collision impulses. To contextualize these challenges within the current state of the art, Table \ref{tab:related_work} provides an analysis of recent co-simulation and multi-agent frameworks developed within the \ac{CARLA} ecosystem.

\begin{table*}[tbp]
	\caption{Comparison of \ac{CARLA} Co-Simulation and Multi-Agent Frameworks.}
	\label{tab:related_work}
	\centering
	\small % Reduces font size slightly to fit the page better
	\renewcommand{\arraystretch}{1.2} % Slightly reduced from 1.3 to save vertical space
	\setlength{\tabcolsep}{3pt} % Tighter horizontal spacing between columns
	
	% Math check: 4 'L' columns. 0.9 + 0.9 + 0.8 + 1.4 = 4.0.
	\begin{tabularx}{\textwidth}{>{\hsize=0.9\hsize}L >{\hsize=0.9\hsize}L >{\hsize=0.8\hsize}L c c c c >{\hsize=1.4\hsize}L}
		\toprule
		\textbf{Framework} & \textbf{Primary Focus} & \textbf{Tech Stack} & \textbf{Dist. Physics} & \textbf{Drive} & \textbf{Bike} & \textbf{Ped.} & \textbf{Key Limitations for \ac{HITL}} \\
		\midrule
		
		DReyeVR \cite{silveraDReyeVRDemocratizingVirtual2022} (2022) & Single-Ego VR / Behavior & \ac{CARLA} fork + Unreal Engine / \ac{VR} & No & $\checkmark$ & -- & -- & Designed primarily for single-node VR; does not describe support for distributed multi-agent haptic interactions. \\
		
		JOAN (2023) \cite{beckersJOANFrameworkHumanautomated2023} & Single-Ego \ac{HITL} \& Human Factors & \ac{CARLA} (Native \ac{API} / \ac{USB}) & No & $\checkmark$ & -- & -- & Focuses on single-node setups; does not describe support for distributed multi-agent deployment across wide-area networks. \\
		
		GrokWalks (2024) \cite{deyGrokWalksPortableVirtual2024} & VRU, Driver-Pedestrian Interaction & CARLA + Mozilla Hubs (WebRTC) & No & $\checkmark$ & -- & $\checkmark$ & WebRTC architecture optimizes for visual \ac{XR} sync, but is not primarily designed for the high-frequency physical coupling required for rigid-body haptics. \\
		
		CDA-HDS (2025) \cite{starkCooperativeDrivingAutomation2025, hanParallelDevelopmentTesting2024} & Highway \ac{V2X} \& \ac{SIL}/\ac{HITL} & CARMA/\ac{CARLA} (\ac{ROS}, \ac{TENA} \cite{noseworthyTestTrainingEnabling2008a}) & No & $\dagger$ & -- & -- & Reliance on high-level \ac{TENA} middleware may introduce serialization overhead, presenting challenges for closed-loop haptic synchronization. \\
		
		Sky-Drive (2025) \cite{huangSkyDriveDistributedMultiAgent2025, skydrive_website} & Socially-aware Multi-agent & \ac{CARLA} (\ac{ROS} / Custom Sync) & No & $\checkmark$ & -- & $\checkmark$ & Centralized physics over standard network middleware is not primarily designed for low-latency remote haptic control loops. \\
		
		\ac{CARMEn} (2026) \cite{machadoCARMEnCARlaBasedMultiAgent2026} & Multi-Agent Immersive XR & \ac{CARLA} + \ac{UE} Multiplayer (Client-Server) & No & $\checkmark$ & -- & $\checkmark$ & Standard \ac{UE} client-server model centralizes physics; high \ac{XR} and audio bandwidth may introduce latency that affects real-time haptic feedback. \\
		
		DMAVA (2026) \cite{islamDMAVADistributedMultiAutonomous2026} & Distributed Multi-AV Stacks & AWSIM / \ac{CARLA} (\ac{ROS}\,2, Zenoh) & No & -- & -- & -- & Distributes \ac{AV} software stacks rather than physics calculation; middleware layers may introduce overhead for high-frequency haptic control loops. \\
		
		\midrule
		\rowcolor[gray]{0.9}
		\textbf{CARLAverse (Ours)} & \textbf{Multimodal \ac{HITL} Interaction} & \textbf{\ac{CARLA} (Direct \ac{API} / \ac{RPC})} & \textbf{Yes} & \textbf{$\checkmark$} & \textbf{$\checkmark$} & \textbf{$\checkmark$} & \textbf{Requires high-performance simulation computer at local node for rendering and local physics} \\
		\bottomrule
	\end{tabularx}
	\par\vspace{1mm}
	\footnotesize
	\textit{Note:} $\checkmark$ = explicitly documented human-operated \ac{HITL} node;
	$\times$ = explicitly unsupported;
	-- = no evidence for or against identified;
	$\dagger$ = \ac{HITL} capability specified or targeted, but no implemented node verified.
	Symbols refer to human-operated nodes, not \ac{NPC} capabilities.
	\textit{Key Limitations for \ac{HITL}} are source-reported limitations or conservative inferences from undocumented capabilities; they do not imply that a framework cannot be extended. \textit{Dist. Physics} refers to distributed local ego-physics for latency-critical human control and haptic feedback.
\end{table*}

As synthesized in the table, frameworks such as \ac{DMAVA} \cite{islamDMAVADistributedMultiAutonomous2026}, \ac{CDA-HDS} \cite{hanParallelDevelopmentTesting2024, starkCooperativeDrivingAutomation2025}, and Sky-Drive \cite{huangSkyDriveDistributedMultiAgent2025} successfully scale autonomous agents, typically by layering communication middleware (e.g., \ac{ROS}\,2, Zenoh, \ac{TENA}) over centralized physics engines. Transmitting high-frequency human control inputs over middleware to a central server may introduce latency overhead for haptic multi-agent research. Similarly, while GrokWalks \cite{deyGrokWalksPortableVirtual2024} excels in visual \ac{VRU} synchronization, and CARJAN \cite{neisCARJANAgentBasedGeneration2025} introduces semantic reasoning, these platforms are not primarily designed for coupled high-frequency force-feedback across \acp{WAN}.

\subsection{Synthesis: The CARLAverse Differentiation}
\label{subsec:carlaverse_diff}

The literature indicates that centralized physics architectures relying on intermediary middleware may introduce latency and jitter challenges when supporting high-frequency, globally connected force-feedback simulators. The CARLAverse ecosystem is engineered specifically to address these structural trade-offs. By bypassing middleware translation layers for active control in favor of native orchestration, the CARLAverse distinguishes itself through three foundational pillars.

Unlike co-simulations reliant on \ac{AI}-driven \acp{NPC}, the CARLAverse integrates multiple human participants into a shared, synchronous, and multimodal \ac{HITL} system. It provides specialized, open-source nodes for vehicles (DrivoCARLA), physical bicycles connected via \ac{BLE} trainers (CycloCARLA), and pedestrians tracked via OpenXR \cite{khronos_openxr} (WalkoCARLA), enabling researchers to capture authentic human-human interactions in mixed traffic. Standard simulators calculate both ego-vehicle and environmental physics on a single machine, creating a severe bottleneck for multi-agent expansion. The CARLAverse implements a decoupled, distributed physics architecture. Latency-critical ego physics (e.g., tire friction, gyroscopic forces, and force feedback) are calculated locally on dedicated client local nodes. Concurrently, the centrally coordinated global traffic and collision matrices are managed by a central server. By localizing high-frequency control loops, the architecture inherently tolerates \ac{WAN} latency, facilitating cross-institutional simulator connections. To circumvent the latency constraints of \ac{ROS} and \ac{DDS} in distributed control loops, the CARLAverse uses direct \ac{CARLA} \ac{API} \acp{RPC}, which is designed to reduce latency between human input and virtual execution. Middleware like \ac{ROS} is strategically delegated to background data logging (e.g., rosbags), ensuring robust data acquisition without compromising the haptic fidelity of the multimodal simulation nodes.

\section{CARLAverse Architecture}
\label{sec:carlaverse}

To facilitate a high-fidelity, multi-agent mixed traffic environment without compromising the latency required for \ac{HITL} immersion, the CARLAverse relies on a specialized, distributed software and hardware architecture.

\subsection{System Architecture Overview}

The CARLAverse ecosystem's core topology follows \ac{CARLA}'s strict server-client paradigm, but is extended to support multiple, heterogeneous simulator clients simultaneously. At the center of the architecture is a dedicated central server running the \ac{CARLA} Unreal Engine environment. Various client machines are connected to this central server via a network. Each local node hosts a specific multimodal simulator setup, such as a DrivoCARLA driving simulator, a CycloCARLA bicycle simulator, or a WalkoCARLA pedestrian setup.

\begin{figure*}[tbp]
	\centering
	\includegraphics[width=0.8\textwidth]{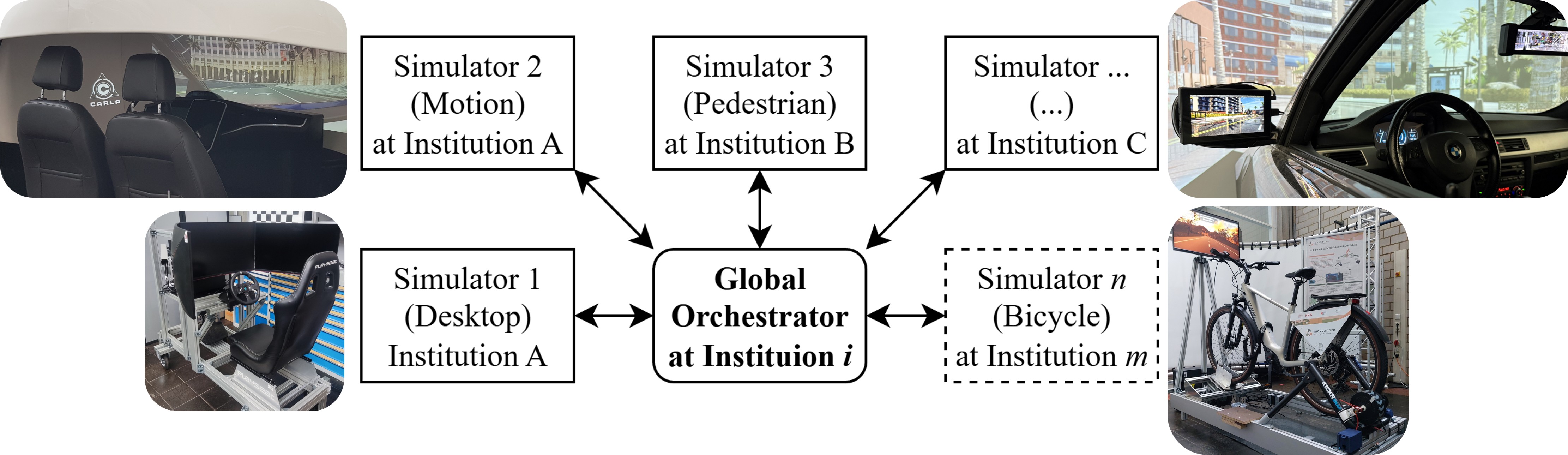} 
	\caption{The CARLAverse multi-server topology. A central global orchestrator synchronizes heterogeneous simulator nodes (e.g., desktop, motion, bicycle, and pedestrian setups) that can be geographically distributed across multiple cooperating research institutions.}
	\label{fig:concept}
\end{figure*}

Crucially, this topology overcomes the physical limitations of a single laboratory. As shown in Figure~\ref{fig:concept}, the CARLAverse allows for geographically distributed, cross-institutional co-simulation. For instance, researchers at Institution A can operate a desktop and a high-fidelity vehicle simulator while participants at Institutions B and C interact simultaneously as cyclists or pedestrians within an identical virtual intersection via a \ac{WAN}. While this interconnected configuration unlocks unprecedented possibilities for large-scale, naturalistic, mixed-traffic research, routing real-time simulator data across vast physical distances presents severe latency and synchronization challenges. To overcome these limitations, the CARLAverse employs a strictly distributed physics architecture.

\subsection{Distributed Physics Calculation}

\begin{figure*}[tbp]
	\centering
	\includegraphics[width=0.9\textwidth]{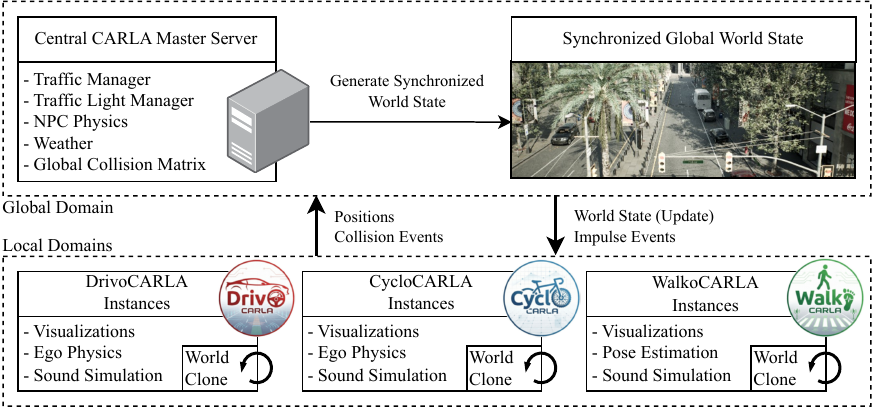} 
	\caption{The distributed physics architecture of the CARLAverse. The central server handles the global domain (centrally coordinated traffic and collisions), while local nodes independently compute latency-critical ego-physics and pose estimation, exchanging only synchronized world states and impulse events over the network.}
	\label{fig:distributed_physics}
\end{figure*}

The most significant technical challenge in networked multi-agent simulators is managing the computational load while reducing latency for human participants. In the CARLAverse, the architecture addresses the trade-off between centralized traffic management and low-latency local execution by strictly decoupling the physics calculations into global and local domains, as illustrated in Figure \ref{fig:distributed_physics}. To maintain global consistency across the distributed network, the central server acts as an authoritative orchestrator. Rather than relying on a middleware bridge to broadcast the global traffic state, the central server operates in a direct server--client orchestration model. It establishes direct \ac{RPC} client connections to every local node.

The central server is exclusively responsible for computing the centrally coordinated elements of the simulation. This includes the \ac{CARLA} Traffic Manager, the Traffic Light Manager, the behavior and physics of all \acp{NPC} (\ac{AI} vehicles and pedestrians), weather conditions (synchronization across local nodes), and the global collision matrix. By centralizing these tasks, the environment remains globally synchronized and consistent for all connected participants. Conversely, the complex and latency-critical dynamics of the human actors are calculated locally on their respective local nodes. For instance, in DrivoCARLA and CycloCARLA, tire-road interactions, the resulting self-aligning torques for force feedback, and motion cueing algorithms are calculated entirely on local workstations. Similarly, WalkoCARLA processes its vision-based pose estimation and \ac{VR} visualization locally to avoid network latency when transmitting high-resolution images. 

This distributed physics approach supports high-frequency local control loops, which often require update rates of several hundred Hertz to feel realistic to a human operator, without being bottlenecked by network latency or the central server's load. To maintain the synchronized global world state, the local nodes only exchange essential, lightweight telemetry (such as absolute position and rotation updates) and receive discrete collision notifications (impulse events) from the central \ac{CARLA} server in case of a collision with an \ac{NPC} or other ego vehicle. To mitigate the visual impact of transient network instability or bandwidth fluctuations, the framework employs client-side position interpolation for all external actors, including \acp{NPC} and other ego vehicles. This ensures smooth rendering even during minor packet loss. In the event of a severe connection loss or a prolonged timeout between a local node and the central orchestrator, the disconnected instance is automatically despawned from the global simulation. This fail-safe mechanism prevents unpredictable collisions resulting from stale telemetry and maintains the integrity of the scenario for all remaining participants. The architecture is designed to accommodate additional local domains, subject to available server and network resources.

\subsection{Communication Protocols and Data Logging}

In early iterations of modular simulator frameworks, middleware such as the \ac{ROS} was used for orchestrating the various software modules. However, as previously established, routing high-frequency control signals through \ac{ROS} nodes introduces unwanted communication overhead, serialization latency, and \ac{DDS} jitter, which severely degrades the fidelity of mechanical feedback systems. 

\begin{figure*}[tbp]
	\centering
	\includegraphics[width=0.9\textwidth]{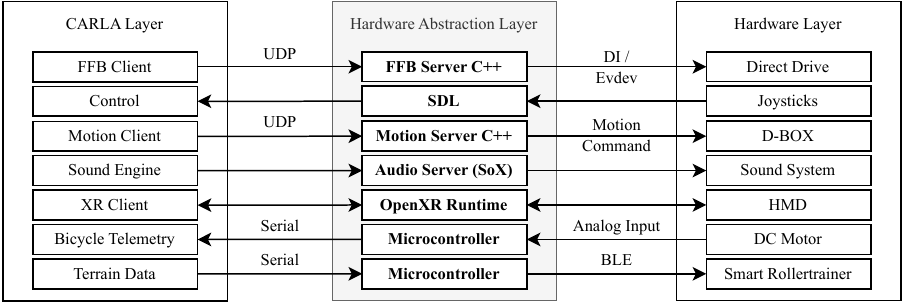} 
	\caption{System architecture of the multimodal \ac{HITL} simulation ecosystem. The framework illustrates the strict decoupling of the high-level simulation core (CARLA Layer) from the physical testbed (Hardware Layer). A dedicated \ac{HAL} acts as specialized middleware, translating network-based simulation telemetry (e.g., \ac{UDP}, Serial) into low-level hardware protocols (e.g., DirectInput (DI)/Evdev, \ac{BLE}, OpenXR). This modular approach concurrently supports diverse feedback modalities, including automotive interfaces (motion platforms, direct drive), immersive audiovisual rendering (\ac{HMD}, SoX), and bicycle-specific physical interfaces (smart rollertrainers, \ac{DC} motors).}
	\label{fig:hal}
\end{figure*}

To eliminate this overhead, the CARLAverse architecture, as illustrated in Figure \ref{fig:hal}, completely bypasses \ac{ROS} for simulator control and orchestration. Instead, all interconnected sub-systems communicate directly using the native \ac{CARLA} \ac{API} via \ac{RPC} over \ac{TCP}/\ac{IP} for reliable transactional state updates. Furthermore, latency-critical hardware telemetry (such as high-frequency force feedback and motion cueing) is routed via dedicated, connectionless \ac{UDP} streams. This decoupling supports high-frequency local control loops by bypassing the transmission overhead of \ac{TCP}. In a traditional centralized architecture, the total haptic loop latency $T_{\text{central}}$ includes significant network components:

\begin{equation} \label{eq:latency_central}
	T_{\text{central}} = t_{\text{input}} + t_{\text{network,tx}} + t_{\text{physics}} + t_{\text{network,rx}} + t_{\text{actuation}}
\end{equation}

In contrast, the CARLAverse's localized physics architecture eliminates the network transmission delays ($t_{\text{network,tx}}$ and $t_{\text{network,rx}}$) from the time-critical loop, resulting in a strictly localized haptic delay $T_{\text{local}}$:

\begin{equation} \label{eq:latency_local}
	T_{\text{local}} = t_{\text{input}} + t_{\text{physics,local}} + t_{\text{actuation}}
\end{equation}

By removing the intermediary publish-subscribe middleware and leveraging these direct, specialized network protocols alongside distributed physics, the architecture is designed to reduce transport delay and promote haptic stability, mitigating the risk of transient network jitter translating into erratic physical force spikes or steering wheel oscillations.

While \ac{ROS} is not used for active control by humans, it still plays a vital role in the ecosystem as a passive, background data-logging engine and interface to automated driving functions (see Figure~\ref{fig:drivocarla_sys}). During a simulation, the local nodes asynchronously publish their state data to \ac{ROS} topics. By using \textit{rosbags}, the CARLAverse can synchronously record vast amounts of experimental data, including human control inputs, eye-tracking coordinates, globally synchronized traffic states, and sensor data, across all connected multimodal simulators simultaneously. Because this publication is handled asynchronously, it enables robust, time-stamped data acquisition for post-experiment analysis without blocking or interfering with the low-latency requirements of the \ac{HITL} physics loop.

\section{Cross-Cutting Frameworks and Orchestration}
\label{sec:crosscut}

Creating a cohesive, shared virtual environment for multiple human actors requires more than just distributed physics. To ensure high modularity and valid experimental results across all nodes, the CARLAverse relies on cross-cutting technologies that manage system orchestration, telemetry visualization, and sensory augmentation uniformly across the framework. In accordance with the system-level focus of this paper, the following subsections outline the software architecture, abstraction layers, and integration interfaces of these cross-cutting subsystems (such as spatial audio, force feedback, and motion cueing), whereas their domain-specific mathematical modeling and validation are addressed in companion works.

\subsection{The CARLAverse Tools Library}

To facilitate rapid development and ensure standardized interactions with the central \ac{CARLA} server, the ecosystem provides a centralized Python library called \texttt{carlaverse\_tools}. This library abstracts repetitive \ac{API} calls and provides shared utility classes used by all multimodal nodes. Furthermore, a \ac{GUI} launcher is provided to make installation, setup and configuration changes more user-friendly.

\begin{figure}[tbp]
	\centering
	\includegraphics[width=\columnwidth]{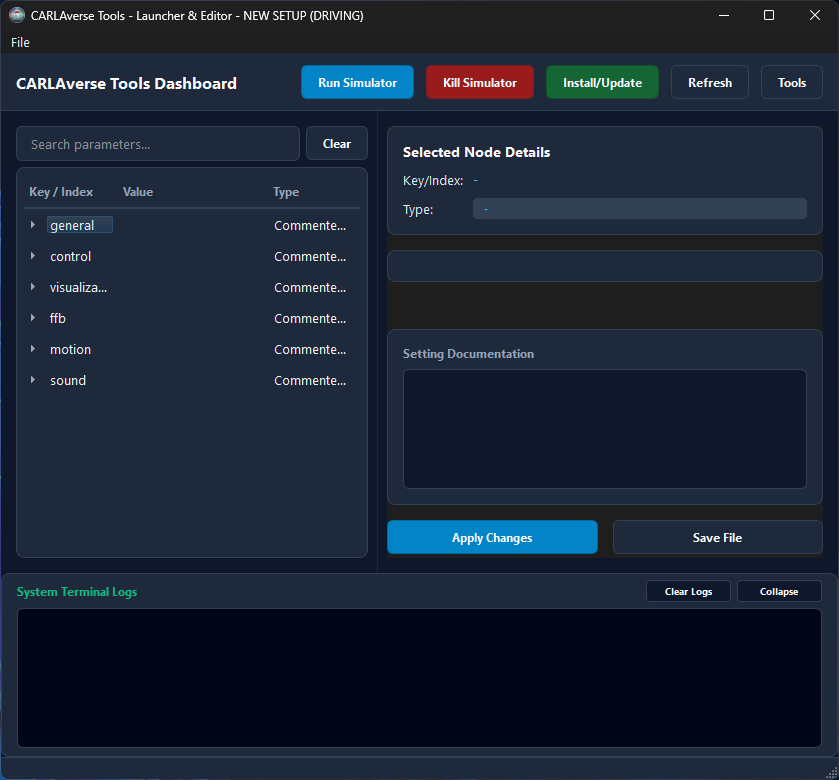} 
\caption{\ac{UI} of the CARLAverse launcher. The dashboard features controls for managing the simulator lifecycle, an integrated tree-view editor for intuitive adjustment of module configurations (e.g., control, \ac{FFB}, motion), and a terminal log panel for real-time system monitoring.}	\label{fig:carlaverse-launcher}
\end{figure}

\begin{itemize}
	\item \textbf{Launcher:} Provides a comprehensive graphical interface for initializing, configuring, and terminating the simulator instances. A streamlined one-click installer facilitates the rapid deployment of the driving, bicycle, and pedestrian simulators. Furthermore, integrated diagnostic tools enable the testing of hardware peripherals, such as \ac{FFB} modules, joystick axes, and buttons (see Figure~\ref{fig:carlaverse-launcher}).
	\item \textbf{ConfigManager:} Manages the loading, parsing, and exporting of the \ac{YAML} configuration files (see Section~\ref{sec:yaml}). It supports dynamic path resolution (e.g., loading from absolute paths or a centralized \texttt{configs} directory) and ensures that all modules instantiate with the correct parameters.
	\item \textbf{SetupManager:} Handles the lifecycle of the virtual environment, including world state management and the safe spawning and destruction of actors across the distributed network.
	\item \textbf{SharedMemory:} Manages interprocess communication between the CARLAverse modules for state variables that must bypass the \ac{CARLA} server.
	\item \textbf{Vehicle Physics and Control:} Provides standardized wrappers to extract latency-critical telemetry (such as velocity, engine RPM, and wheelbase dimensions) and seamlessly manage actuator commands, such as vehicle lighting and turn signals.
	\item \textbf{Utilities:} A suite of helper functions designed to streamline actor and sensor management. This includes fail-safes for retrieving specific actors, ensuring the unique spawning of sensors (preventing duplicate data streams or \ac{API} crashes), and robust error handling during runtime.
\end{itemize}

\subsection{Rapid Prototyping and YAML-Based Orchestration}
\label{sec:yaml}

A frequent limitation of custom simulator software is the tight coupling of hardware mappings, where transitioning from a basic desktop setup to a high-fidelity motion platform necessitates extensive source code modifications. To address this, the CARLAverse adheres to the principle of separation of concerns by implementing a \textit{Single Source of Truth} architecture. Executable modules, such as control, force feedback, and visualization, are fundamentally hardware-agnostic. Instead, the instantiation of each simulator node is governed entirely by a comprehensive \ac{YAML} configuration file, facilitating rapid prototyping and streamlined reconfiguration across multiple domains.

Furthermore, strict experimental reproducibility is a fundamental requirement for behavioral and human factors research. By archiving this central \ac{YAML} configuration file alongside the experimental \texttt{rosbag} data logs, researchers can ensure that the specific hardware state, physics models, environmental profiles, and control mappings are accurately preserved and can be consistently reconstructed long after an experiment concludes.

\subsection{Web-Based HMI and Telemetry Framework}

\begin{figure*}[tbp]
	\centering
	\includegraphics[width=0.9\textwidth]{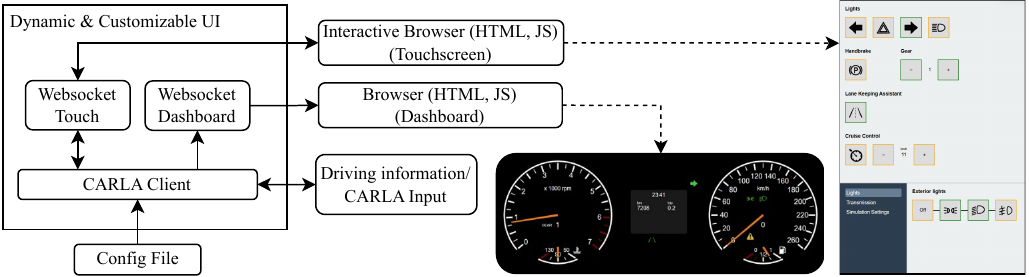} 
	\caption{Architecture of the Dynamic and Customizable \ac{UI}. The diagram illustrates the data flow and system components connecting the \ac{CARLA} Client to web-based user interfaces via WebSockets. Configured via an external file, the system drives two distinct browser displays: an interactive touchscreen for inputting vehicle controls (e.g., lights, gear selection, cruise control) and a dashboard display for visualizing real-time driving information (e.g., speedometer and tachometer).}
	\label{fig:userinterface}
\end{figure*}

Instead of relying on hardcoded Unreal Engine widgets that break during version updates, the CARLAverse employs a decoupled, web-based \ac{HMI} framework. Driven dynamically by the local ego-vehicle's telemetry via \ac{API} callbacks, this framework hosts a local web server to display real-time simulation data as shown in Figure~\ref{fig:userinterface}. 

While initially developed to render dynamic dashboards (speedometer, tachometer, \ac{ADAS} status indicators) for the DrivoCARLA vehicle cabins, its cross-cutting nature allows it to be used universally, especially in CycloCARLA. Researchers can rapidly prototype and overlay new infotainment interfaces, \ac{VRU} warning signals, or external experimenter control panels using standard web technologies (\ac{HTML}/JavaScript) without recompiling the simulator.

\subsection{Spatial Audio Rendering}

In mixed-traffic scenarios, auditory cues are just as critical as visual information. For pedestrians (WalkoCARLA) and cyclists (CycloCARLA), acoustic localization of approaching vehicles is a primary mechanism for situational awareness and risk assessment. For drivers (DrivoCARLA), realistic engine and environmental sounds are essential for speed perception and immersion. Native simulation environments often lack the sophisticated, dynamic sound propagation models required for high-fidelity \ac{HITL} testing. To address this, the CARLAverse integrates a scalable sub module. This module acts as a centralized spatial audio engine that generates a highly realistic, multi-channel soundscape for each connected client based on their absolute position in the virtual world.

\begin{itemize}
	\item \textbf{Geo-Spatial Sound Generation:} The engine extracts metadata from the active OpenDRIVE map and the Unreal Engine environment to generate context-aware ambient noise. For example, it calculates the proximity to virtual vegetation to generate wind rustling, or proximity to water bodies for ocean acoustics.
	\item \textbf{Dynamic Emitters:} Localized events, such as construction noise, sirens, or the tire-road friction of surrounding AI traffic, are spawned as dynamic audio emitters in the 3D space.
	\item \textbf{Client-Specific Rendering:} The central server calculates the spatial coordinates of all sound emitters and transmits them to the respective local nodes. Each simulator node then renders the audio mix locally. This can be done from simple stereo systems (e.g. headsets) to multi-channel.
\end{itemize}

While this geo-spatial sound generation significantly enhances immersion, the detailed acoustic modeling, frequency masking, and subjective validation of the CARLAcoustics module are highly complex and will be addressed in a dedicated future publication.

\subsection{Force Feedback Server}
\label{sec:ffb}

To bridge the gap between the local physics engine and the physical hardware, the CARLAverse uses a custom, standalone C++ \ac{FFB} server. The local physics engine calculates the required steering resistance and counter-torques, continuously streaming these parameters to the C++ server via a non-blocking \ac{UDP} connection. To eliminate memory padding and minimize serialization overhead, the commands are tightly packed into a strict 40-byte \texttt{C++ struct} using 1-byte memory alignment, as detailed in Table~\ref{tab:ffb_message}.

The server parses this incoming 40-byte stream and translates the individual 32-bit integers into either DirectInput (windows) or evdev (linux) commands. This pipeline allows for the simultaneous application of complex haptic layers, including spring offsets for autopilot centering, viscous dampers, mechanical friction for tire scrubbing, and periodic vibrations for micro-terrain surfaces. Because this C++ server is entirely hardware-agnostic at the simulation level, it operates as a shared core module used by both DrivoCARLA (for automotive steering bases) and CycloCARLA (for bicycle front-fork motors). Furthermore, it can also be used in custom simulation environments as a standalone extension.

Crucially, to ensure participant safety during \ac{HITL} experiments, the \ac{FFB} server features a built-in hardware watchdog mechanism. If the simulation crashes or the network connection drops for longer than a predefined timeout threshold, the server automatically engages a safety stop. This routine instantaneously zeros out all active forces (including constant, spring, and damper forces) to prevent the high-torque direct-drive motors from locking up or spinning uncontrollably. Furthermore, the server is configurable via command-line arguments, enabling researchers to quickly assign \ac{UDP} ports, select specific hardware devices, and adjust watchdog timeouts without needing to recompile the C++ source code.

\begin{table}[tbp]
	\caption{\ac{UDP} Payload Structure for \ac{FFB} Commands}
	\label{tab:ffb_message}
	\centering
	\renewcommand{\arraystretch}{1.2} 
	% Changed tabularx to tabular and replaced the 'X' column with an 'l' (left-aligned) column
	\begin{tabularx}{\columnwidth}{c c L}
		\toprule
		\textbf{Byte} & \textbf{Type} & \textbf{Parameter / Description} \\
		\midrule
		0--3 & \texttt{int32\_t} & \textbf{Spring Stiffness:} Strength of the pull towards the target offset (0 to 10000). \\
		4--7 & \texttt{int32\_t} & \textbf{Spring Offset:} Target center position for the spring (-10000 to 10000). \\
		8--11 & \texttt{int32\_t} & \textbf{Spring Deadband:} Zero-force zone around the target to prevent oscillation. \\
		12--15 & \texttt{int32\_t} & \textbf{Spring Saturation:} Maximum force ceiling for safety. \\
		16--19 & \texttt{int32\_t} & \textbf{Constant Force:} Directional sustained force (e.g., G-forces in corners). \\
		20--23 & \texttt{int32\_t} & \textbf{Damper:} Viscous rotation resistance to add weight to movement. \\
		24--27 & \texttt{int32\_t} & \textbf{Friction:} Constant mechanical friction (e.g., tire scrubbing). \\
		28--31 & \texttt{int32\_t} & \textbf{Inertia:} Resistance to acceleration/changes in direction. \\
		32--35 & \texttt{int32\_t} & \textbf{Vibration Magnitude:} Strength of the rumble effect (e.g., engine RPM). \\
		36--39 & \texttt{int32\_t} & \textbf{Vibration Frequency:} Frequency of the vibration cycle in Hz. \\
		\bottomrule
	\end{tabularx}
\end{table}

\subsection{Motion Platform Server}

Similar to the force feedback pipeline, the CARLAverse uses a standalone C++ Motion Platform Server to translate simulation telemetry into physical motion. This server receives real-time spatial parameters from the local physics engine via a low-latency, connectionless \ac{UDP} stream, ensuring zero head-of-line blocking. To maximize performance and minimize serialization overhead, the telemetry commands are tightly packed into a strict 21-byte binary payload using little-endian byte order, as detailed in Table~\ref{tab:motion_message}.

Under the hood, this server parses incoming datagrams and integrates directly with the D-BOX LiveMotion SDK to actuate the motion rig. Because it interfaces specifically with D-BOX hardware, operating this particular server module requires a valid D-BOX software license. However, the system architecture is intentionally designed to be hardware-agnostic at the simulation level. If researchers wish to use a different motion platform, only this C++ backend server needs to be exchanged. The client-side implementation within the physics engine remains completely untouched, as it will continue to broadcast the exact same standard \ac{UDP} binary protocol regardless of the physical hardware in use. Because this motion client already provides pre-processed, washout-filtered telemetry, the custom server only needs to map these commands into hardware-specific instructions.

For continuous motion simulation, the server relies on a \texttt{FRAME\_UPDATE} command (\texttt{0x04}) transmitted at high frequencies. This payload carries five 32-bit floats representing roll, pitch, heave, engine RPM, and torque. To facilitate rapid deployment and testing, the server is configurable via command-line arguments, allowing researchers to assign custom \ac{UDP} ports and toggle verbose logging without needing to recompile the C++ source code. Furthermore, the server features built-in safety checks; if a packet is received that is shorter than expected (e.g., missing floats), a warning is logged and the data is safely ignored to prevent erratic hardware behavior.

\begin{table}[tbp]
	\caption{\ac{UDP} Payload Structure for \texttt{FRAME\_UPDATE} Commands of the Motion Platform Server}
	\label{tab:motion_message}
	\centering
	\renewcommand{\arraystretch}{1.2}
	\begin{tabularx}{\columnwidth}{c c L}
		\toprule
		\textbf{Byte} & \textbf{Type} & \textbf{Parameter / Description} \\
		\midrule
		0 & \texttt{uint8\_t} & \textbf{Command ID:} Identifier for the action (e.g., \texttt{0x04} for \texttt{FRAME\_UPDATE}). \\
		1--4 & \texttt{float} & \textbf{Roll:} Kinematic roll value of the simulated vehicle. \\
		5--8 & \texttt{float} & \textbf{Pitch:} Kinematic pitch value of the simulated vehicle. \\
		9--12 & \texttt{float} & \textbf{Heave:} Vertical displacement (heave) of the simulated vehicle. \\
		13--16 & \texttt{float} & \textbf{Engine RPM:} Current engine revolutions per minute for haptic vibration. \\
		17--20 & \texttt{float} & \textbf{Engine Torque:} Current engine torque output for dynamic haptic effects. \\
		\bottomrule
	\end{tabularx}
\end{table}

\subsection{Visualization}

\begin{figure*}[tbp]
	\centering
	\includegraphics[width=0.85\textwidth]{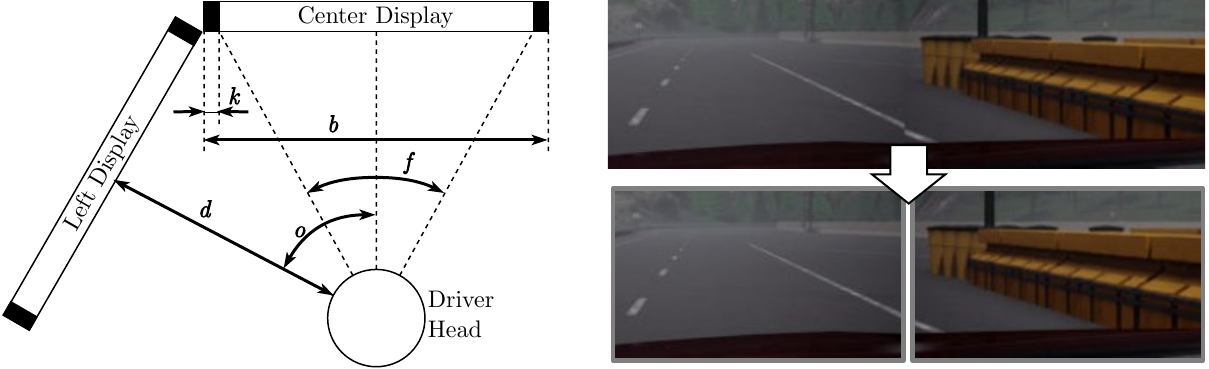} 
	\caption{Example of offset compensation in physical mode for simulators with displays. An automatically calculated offset based on the user-defined parameters resulting in a seamless image on the custom hardware setup.} 
	\label{fig:displays}
\end{figure*}

The visualization module is shared across DrivoCARLA, CycloCARLA, and WalkoCARLA (when not operating in \ac{VR} mode). Upon initialization, camera sensors are attached to the simulated vehicle or pedestrian. There are three distinct modes for camera configuration: \textit{simple}, \textit{physical}, and \textit{manual}. While the simple mode assumes a single continuous screen where multiple camera views can be projected for parallel rendering, the manual mode enables custom setups, such as a \ac{CAVE} environment or specific monitor mountings. In the physical mode, the number of cameras $M$ is equal to the number of monitors or video projectors used in the simulator. This value is specified in the configuration file which also contains information about other parameters, such as the resolution, the width of the monitor $b$, the width of the side monitor frame $k$, and the distance between the screens and the driver's head $d$ as described in Figure~\ref{fig:displays}. The visualization module uses these values to calculate the horizontal \ac{FOV} $f$ for each monitor and the horizontal rotation angles $o_i$ of the camera sensors, ensuring that the simulated scene is displayed in a realistic manner.

\begin{equation} \label{eq:fov}
	f = 2 \cdot \arctan\left(\frac{\left(\frac{b}{2} - k\right)}{d}\right)
\end{equation}

where $b$ is the width of a single monitor, $d$ is the distance from the driver's head to the monitor surface, and $k$ is defined by the frame width of the monitor. While the horizontal \ac{FOV} $f$ remains constant for all camera sensors with identical monitors, the rotation angles $o_i$ of these cameras vary by a factor $i$, where $i$ is calculated based on whether the number of displays $M$ is an even or odd integer. Thus, $i$ represents a list of numbers for the partial rotation of each monitor:

\begin{equation} \label{eq:yaw}
	\begin{gathered}
		o_i = i \cdot 2 \cdot \arccos\left(\frac{d}{\sqrt{\left(\frac{b}{2}\right)^2 + d^2}}\right) \quad \textrm{with} \\
		i = \begin{cases} \pm n \; \textrm{and} \, n \in \mathbb{N}_0, n \leq \frac{M-1}{2}, &  M \bmod 2 = 1 
			\\  \pm \frac{n}{2} \; \textrm{and} \, n \in \mathbb{N}, n \leq \frac{M}{2}, & M \bmod 2 = 0 \end{cases} 
	\end{gathered}
\end{equation} 

For example, three monitors result in $i = \{-1, 0, 1\}$. The position of the camera sensors is identical to the position of the driver's head in the vehicle coordinate system, as specified in the configuration file. They are then displayed side by side with a total resolution of the screens as shown in Figure~\ref{fig:displays}. Taking into account the monitor frames, rotations and distances, the image from the camera sensors will appear smooth and without shifting on the driver's monitor.

\section{Multimodal Simulation Nodes in the CARLAverse}
\label{sec:simdetail}

The theoretical architecture of distributed physics and direct \ac{API} communication forms the backbone of the CARLAverse. However, the physical manifestation of this ecosystem consists of specialized, multimodal simulator nodes. Each node acts as a highly immersive client designed to replicate the specific perceptual cues (visual, auditory, haptic, and vestibular) required for a particular type of road user. The ecosystem currently encompasses three primary domains: DrivoCARLA (vehicles), CycloCARLA (bicycles), and WalkoCARLA (pedestrians). The descriptions in this section document the integration interfaces, hardware abstractions, and operational roles of each simulator node within the connected ecosystem (including bicycle dynamics, motion cueing, \ac{VR} rendering, and pose estimation); standalone model-level validation of these individual nodes is reserved for companion publications.

\subsection{DrivoCARLA: Vehicle Simulator}

\begin{figure}[tbp]
	\centering
	\includegraphics[width=\columnwidth]{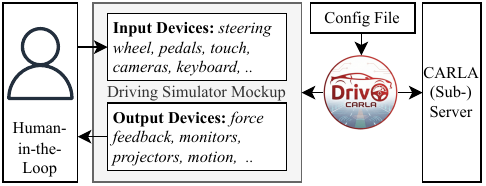} 
	\caption{System architecture of the DrivoCARLA framework. DrivoCARLA serves as the central orchestrator for the \ac{HITL} driving simulator, directly bridging human input devices (e.g., steering wheels, pedals, and cameras) with the \ac{CARLA} server. By managing this closed-loop communication, DrivoCARLA translates real-time simulation data into immersive visual rendering and force feedback haptics for the driver.}
	\label{fig:drivocarla_overview}
\end{figure}

DrivoCARLA serves as the flagship automotive node within the CARLAverse. It is designed as a scalable framework capable of adapting to any level of simulator immersion, ranging from simple single-monitor desktop setups to fully enclosed, multi-display vehicle cabin mock-ups (see Figure~\ref{fig:drivocarla_overview}). Because cross-cutting systems like visualization, spatial audio, and haptic \ac{FFB} interfaces are provided globally by the ecosystem, this section of the DrivoCARLA node focuses entirely on vehicle control, hardware abstraction, and automotive systems integration.

\begin{figure*}[tbp]
	\centering
	\includegraphics[width=\textwidth]{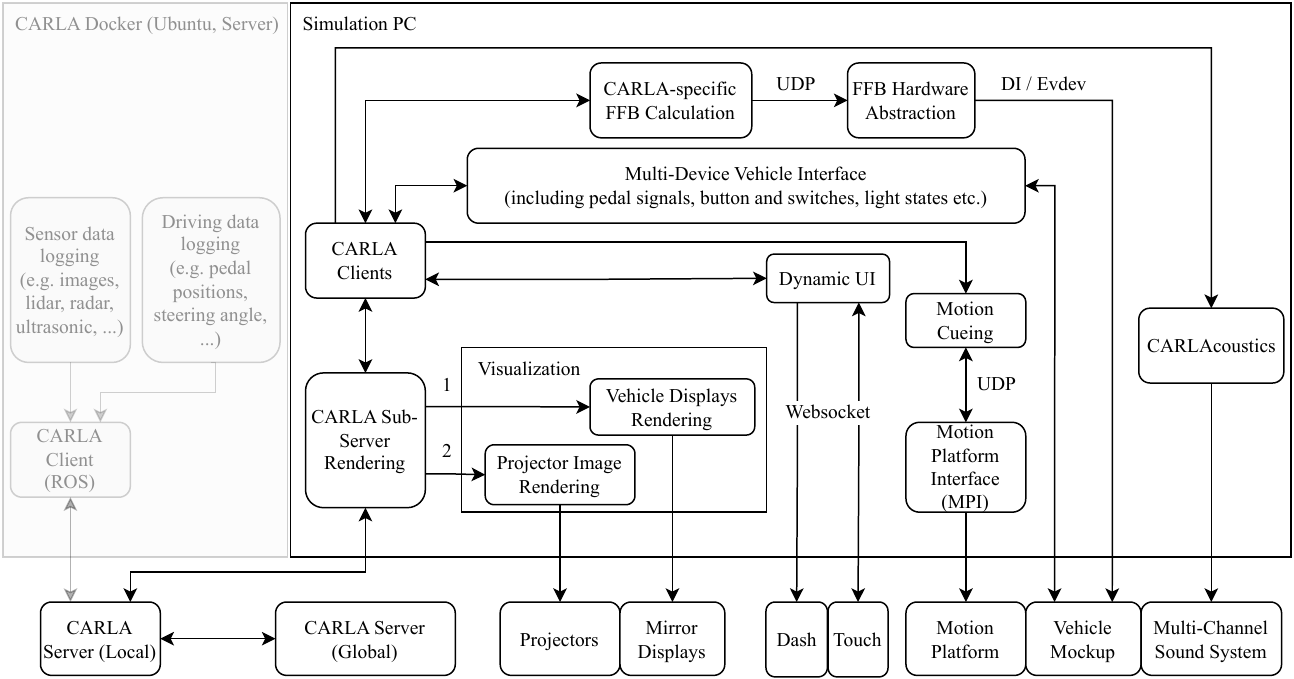} 
	\caption{Detailed system architecture of the DrivoCARLA node. The diagram illustrates the data flow between the central \ac{CARLA} server and the local simulation PC, highlighting the integration of visualization, dynamic \ac{UI}, motion cueing, and multi-device vehicle interfaces. It also depicts the asynchronous \ac{ROS}-based data logging pipeline for sensor and telemetry data.}
	\label{fig:drivocarla_sys}
\end{figure*}

To orchestrate components, DrivoCARLA employs a comprehensive internal software architecture, as illustrated in Figure~\ref{fig:drivocarla_sys}. The local Simulation PC hosts dedicated modules for multi-device vehicle interfaces, dynamic \ac{UI} generation, motion cueing, and hardware-specific abstractions such as force feedback. It communicates bidirectionally with the central \ac{CARLA} server while optionally routing sensor and driving telemetry to an asynchronous \ac{ROS}-based logging container, supporting robust data capture without interfering with the real-time haptic control loops. An early iteration of this pipeline has already demonstrated its empirical utility and standalone viability in a preliminary user study \cite{reblingMindGapQuantifying2026}.

\subsubsection{Control and ADAS Interfaces}

The control module manages all human inputs, abstracting diverse physical hardware (steering wheels, pedal boxes, and button arrays) via the platform-independent Pygame library \cite{pygame}. As defined in the central \ac{YAML} configuration, these raw, linear hardware inputs ($-1 \le x \le 1$) are mathematically transformed into non-linear vehicle commands ($y$) through custom curves before being sent to the local physics engine. To maximize flexibility, DrivoCARLA does not hardcode these transformations. Instead, the system reads arbitrary mathematical string expressions directly from the \ac{YAML} payload (e.g., \texttt{math.pow(x, 2)}) and dynamically compile them into executable lambda functions at runtime. This allows researchers to define completely customized control mappings without modifying the underlying source code. For instance, to map brake pressure or non-linear steering ratios, an exponential sensitivity curve is commonly applied:

\begin{equation} \label{eq:inputmapping}
	y = \text{sgn}(x) \cdot |x|^\gamma
\end{equation}

where a factor of $\gamma > 1$ flattens the response near the center and increases sensitivity at the extremes. To ensure realistic automotive logic, DrivoCARLA uses complex internal state machines; for instance, ensuring that turn signals override hazard lights temporarily as described below. 

Furthermore, the control pipeline exposes dedicated software interfaces to facilitate \ac{HITL} testing of \ac{ADAS}. External control algorithms can inject \ac{API} override commands directly into the data stream. For example, a \ac{CC} algorithm can overwrite the driver's pedal inputs, or a \ac{LKAS} can inject counter-torques directly into the C++ \ac{FFB} server, allowing researchers to study human trust and take-over reaction times during automated interventions.

\begin{figure*}[tbp]
	\includegraphics[width=\textwidth]{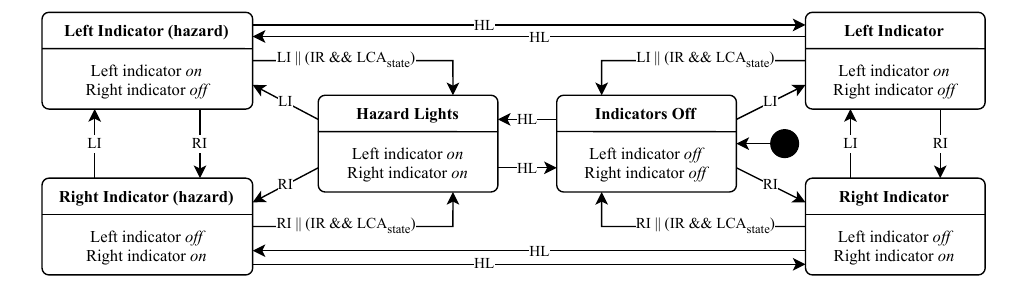} 
	\caption{Indicator System State Machine. State transitions are triggered by rising edge events: Hazard Light (HL), Left Indicator (LI), Right Indicator (RI), and Indicator Reset (IR). The diagram illustrates that directional resets via the IR event are conditional and only execute if the Lane Change Assist ($\text{LCA}_\text{state}$) is active.}
	\label{fig:drivocarla_indicator}
\end{figure*}

Gears can be shifted manually by the driver or automatically. While brake lights and rear lights are activated and deactivated according to the vehicle's operating status, most other lights are controlled by the driver, such as high beams or interior lighting. State machines are used to ensure that the functionality of the indicators is consistent with real-world expectations, and to control the switching between parking, low beam, and fog lights. By way of illustration, the state machine for the indicators is shown in Figure~\ref{fig:drivocarla_indicator}. It can be seen, for example, that the left-hand indicator is deactivated when the right-hand indicator is activated, and that the hazard warning lights can be temporarily deactivated using the standard indicator switches, but are reactivated when the indicator is deactivated.

\subsubsection{Steering Physics and Force Feedback}

To deliver highly realistic steering resistance, DrivoCARLA's local physics engine computes the haptic feedback dynamically before streaming the torque values to the \ac{FFB} server. Rather than relying on simple linear springs, the engine calculates realistic self-aligning torque $M_z$ at the tire contact patch using a simplified Pacejka Magic Formula based on the current slip angle $\alpha$, vertical tire load $F_z$, and pneumatic trail $t_p$:

\begin{equation} \label{eq:pacejka}
	M_z = -F_z \cdot t_p \cdot \sin(1.6 \cdot \arctan(8.0 \cdot \alpha))
\end{equation}

Because slip-angle mathematics become physically unstable and singular at near-zero velocities, the physics engine implements a speed-dependent safety blending mechanism. Below a threshold of 15\,km/h, the dynamic Pacejka physics are smoothly faded out and replaced by a synthetic, high-resistance parking spring and friction force. This ensures mathematical stability during urban maneuvers like parallel parking or stopping at crosswalks, preventing erratic steering wheel oscillations.

\subsubsection{Motion Cueing}

\begin{figure}[tbp]
	\centering
	\includegraphics[width=0.45\textwidth]{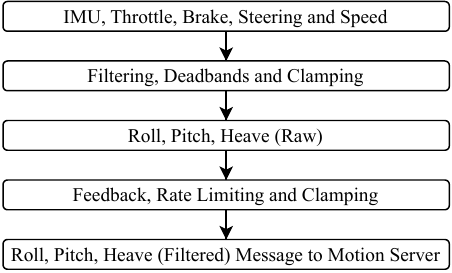} 
	\caption{Signal processing pipeline of the Motion Cueing Algorithm, illustrating the transformation of raw CARLA telemetry into clamped, filtered, and washout-regulated Pitch, Roll, and Heave commands.}
	\label{fig:motion}
\end{figure}

To enhance physical immersion beyond force feedback, DrivoCARLA integrates a signal-based \ac{MCA} to actuate a 3-\ac{DOF} (Pitch, Roll, Heave) motion platform (see Figure~\ref{fig:motion}). Since a reduced \ac{DOF} system cannot physically reproduce the full range of real-world vehicle movements, the \ac{MCA} focuses on a perception-oriented representation of the vehicle dynamics extracted from \ac{CARLA}'s \ac{IMU} and vehicle telemetry. 

The algorithm maps specific driving states to the available platform axes through a combination of continuous and transient movement components. To ensure stability and prevent the motion platform from reaching its mechanical limits, the raw telemetry data must be pre-processed. As a foundational step, all incoming accelerations $a_{x}$, $a_{y}$, and $a_{z}$ are smoothed using a first-order \ac{LPF} to reduce high-frequency noise. 

The Pitch angle $\theta$ is primarily derived from the low-pass filtered longitudinal acceleration $a_{x, \text{LPF}}$, normalized by a reference acceleration $a_{\text{ref}}$. To prevent unrealistic oscillations at a standstill, the acceleration component is scaled by a speed-dependent function $f_{\text{scale}}(v)$ that suppresses the signal at low speeds. Furthermore, the pitch calculation incorporates the brake pedal input as both a continuous proportion and a transient impulse derived from its rate of change to emphasize hard stops:
\begin{equation}
	\theta = \frac{a_{x, \text{LPF}}}{a_{\text{ref}}} \cdot f_{\text{scale}}(v) + k_b \cdot b_{\text{LPF}} + k_{\text{impulse}} \cdot \dot{b}
\end{equation}
where $b_{\text{LPF}}$ is the low-pass filtered brake pedal input, $\dot{b}$ is its rate of change, and $k_b$ and $k_{\text{impulse}}$ are their respective tuning gains.

The Roll angle $\phi$ is driven by a combination of the filtered lateral acceleration $a_{y, \text{LPF}}$, normalized by a reference lateral acceleration $a_{y, \text{ref}}$, and the low-pass filtered steering input $s_{\text{LPF}}$ scaled by a tuning gain $k_s$. The inclusion of the steering signal allows the platform to react to steering maneuvers proactively, before lateral forces have fully accumulated:
\begin{equation}
	\phi = \frac{a_{y, \text{LPF}}}{a_{y, \text{ref}}} + k_s \cdot s_{\text{LPF}}
\end{equation}

Finally, the Heave axis $z$ simulates vertical dynamics, such as driving over curbs or road bumps. Instead of relying on constant vertical states, the Heave is generated by evaluating the dynamic changes, or jerk $j_x = \dot{a}_{x, \text{LPF}}$ and $j_z = \dot{a}_{z, \text{LPF}}$. These jerk signals are processed through a deadband to suppress minor noise, clamped to a maximum value, and low-pass filtered to create smooth vertical impulses:
\begin{equation}
	z = k_{jx} \cdot \widetilde{j}_{x} + k_{jz} \cdot \widetilde{j}_{z} + z_{\text{brake}}
\end{equation}
where $\widetilde{j}_{x}$ and $\widetilde{j}_{z}$ represent the conditioned jerk signals after applying the deadband, clamping, and low-pass filtering, while $k_{jx}$ and $k_{jz}$ are their respective scaling factors. The term $z_{\text{brake}}$ represents an additional negative heave impulse triggered exclusively during emergency braking.

To conclude the motion cycle, washout filters continuously and smoothly return the platform axes to their neutral positions after an acceleration event. The rate of this return is dynamically adapted based on the vehicle's speed, ensuring the motion cues remain plausible and comfortable for the driver.

\subsection{CycloCARLA: Bicycle Simulator}

\begin{figure}[tbp]
	\centering
	\includegraphics[width=\columnwidth]{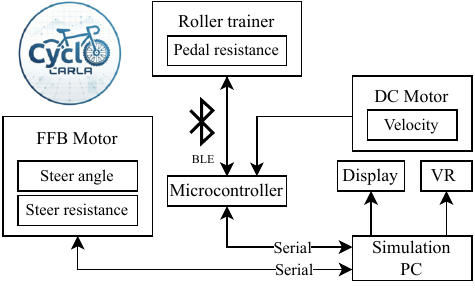} 
	\caption{Hardware and data flow architecture of the CycloCARLA simulator node. A microcontroller (e.g. ESP32) processes high-frequency velocity telemetry to bypass standard \ac{BLE} bottlenecks, while a DirectInput serial connection handles the latency-critical steering force feedback.}
	\label{fig:cyclocarla}
\end{figure}

Bicycles represent a critical class of \acp{VRU}. The kinematics and control strategies of a bicycle are fundamentally different from those of a four-wheeled vehicle. Because \ac{CARLA}'s native physics engine is inherently built around four-wheeled Ackermann steering, CycloCARLA actively overrides the default vehicle dynamics. Instead, it calculates a custom single-track kinematic bicycle model locally and enforces the resulting movement by directly injecting absolute target velocity and angular velocity vectors into the simulation every frame. CycloCARLA is designed to capture these interactions within a stationary \ac{HITL} setup using a physical bicycle frame mounted on a smart resistance roller trainer. The data flow and hardware abstraction, as illustrated in Figure \ref{fig:cyclocarla}, are divided into three primary components: bicycle speed capture via a \ac{DC} motor, pedal resistance adaptation via \ac{BLE}, and steering angle sensing via \ac{SDL} and force feedback via DirectInput/Evdev. The comprehensive mathematical validation of the single-track vehicle dynamics and force feedback loops implemented in CycloCARLA falls outside the scope of this architectural overview and is presented in a dedicated companion study \cite{reblingCycloCARLADesignImplementation2026} (to appear).

\subsubsection{Propulsion and Resistance}

The user's pedaling power is captured to drive the virtual bicycle. Because commercial smart roller trainers typically broadcast telemetry at a low frequency (e.g., 1\,Hz), this latency is insufficient for real-time physics. To overcome this, CycloCARLA uses a microcontroller to capture high-frequency wheel speed data, which is designed to reduce latency in longitudinal acceleration within the \ac{CARLA} world. Simultaneously, the local physics engine calculates the physical forces acting on the virtual bicycle \cite{rauchApproachHolisticModelling2023}, computing the combined longitudinal resistance $F_{\text{res}}$ as the sum of rolling resistance $F_\text{roll}$ with friction coefficient $f_\text{R}$, resistance due to road slope $F_\text{slope}\left(\theta\right)$, and aerodynamic drag $F_\text{drag}$:

\begin{equation} \label{eq:cycloresistance}
\begin{split}
	F_{\text{res}} &= \underbrace{f_\text{R} m g \cos(\theta)}_{\textstyle F_\text{roll}} + \underbrace{m g \sin(\theta)}_{\textstyle F_\text{slope}} \\
	&\quad + \underbrace{\frac{1}{2} \rho_\text{air} c_\text{W} A \left(v_\text{bike}-v_\text{wind}\right)^2}_{\textstyle F_\text{drag}}
\end{split}
\end{equation}

Based on this calculation, the system must translate the total physical force $F_{\text{res}}$ into a resistance percentage (0-100\%) expected by the proprietary smart trainer hardware. To bridge this gap between pure physics and closed-source hardware constraints, CycloCARLA uses a calibrated 2D \ac{LUT}. The system applies bilinear interpolation across the \ac{LUT}, cross-referencing the current forward velocity and the computed $F_{\text{res}}$ to dynamically generate a smooth, fractional resistance percentage. This mapped command is then sent back to the trainer over \ac{BLE} to physically simulate the gradients and environment with high fidelity.

\subsubsection{Steering and Haptics}

Since physical leaning is constrained in the stationary frame, lateral control depends entirely on the steering interface. The front fork is connected to a commercial \ac{FFB} motor that is controlled via the CARLAverse Force Feedback Server (see Section~\ref{sec:ffb}). The local physics engine computes trail-induced self-centering forces and sends constant force commands to the motor, providing realistic steering resistance. Additionally, the interface enables the addition of high-frequency vibration effects, which simulate road surface textures and sudden collision impulses (e.g., caused by curbs).

\subsubsection{Visual Immersion}

Although \ac{VR} headsets can provide a 360-degree field of view for over-the-shoulder checks, stationary bicycle setups often cause visual-vestibular conflict and simulator sickness when using \ac{VR}. Therefore, CycloCARLA's primary visualization uses a monitor positioned directly in front of the test subject to ensure comfort during prolonged studies. However, the system's modular architecture allows researchers to seamlessly switch to \ac{VR} rendering based on the principle of WalkoCARLA or add additional monitors if a specific experimental design strictly requires peripheral immersion.

\subsection{WalkoCARLA: Pedestrian Simulator}

\begin{figure}[tbp]
	\centering
	\includegraphics[width=\columnwidth]{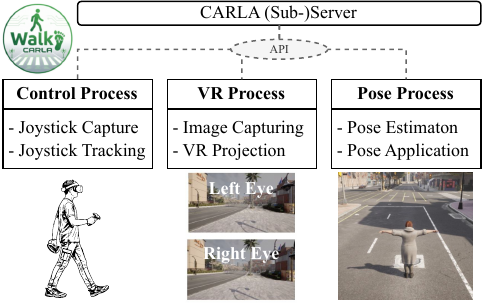}
	\caption{Software architecture of the WalkoCARLA pedestrian node. By strictly using the native \ac{CARLA} \ac{API} to manage the Control, \ac{VR}, and Pose processes, the system avoids hardcoded engine modifications, supporting compatibility with future \ac{CARLA} updates.}
	\label{fig:walkocarla}
\end{figure}

Pedestrians represent the most unpredictable and vulnerable elements in urban mixed traffic. Including them in \ac{HITL} simulations is important for validating crosswalk yielding behaviors and intent recognition algorithms. WalkoCARLA is a \ac{VR}-based pedestrian node. Unlike vehicles or bicycles, pedestrians do not possess a fixed mechanical interface. Therefore, WalkoCARLA relies entirely on spatial tracking and bodily representation. To ensure future compatibility and hardware abstraction, WalkoCARLA uses a strict \textit{\ac{API}-only} software architecture, just as DrivoCARLA and CycloCARLA do. This is illustrated in Figure \ref{fig:walkocarla}. By interacting with the local \ac{CARLA} instance exclusively through the native Python \ac{API}, rather than relying on hardcoded modifications to the Unreal Engine source code, the node remains modular and can be easily migrated to newer \ac{CARLA} releases. The architecture is divided into three parallel processes.

\subsubsection{Control Process}

While natural walking can be mapped one-to-one in the physical laboratory space, traversing larger virtual intersections requires artificial locomotion. The Control Process handles joystick capture and tracking via standard \ac{VR} controllers. To maintain hardware abstraction and avoid dependency on proprietary manufacturer SDKs, WalkoCARLA uses standard OpenXR Action Bindings. These 2D hardware vectors are read dynamically and mapped directly to absolute direction vectors based on the pedestrian's current forward orientation in the world. This allows the user to navigate the expansive \ac{CARLA} environment fluidly without encountering physical room-scale boundaries, regardless of the specific \ac{VR} hardware used. Future work will include implementing omnidirectional treadmills to provide a more immersive control experience.

\subsubsection{VR Process}

To maximize visual fidelity and prevent simulator sickness, the \ac{VR} Process manages the dual-camera image capturing and \ac{VR} projection (Left Eye / Right Eye). Because modern \ac{VR} headsets demand extremely high display resolutions to maintain immersion, rendering two distinct stereoscopic viewpoints simultaneously represents a massive computational bottleneck. To sustain the high frame rates required for valid \ac{HITL} research, the WalkoCARLA architecture is optimized for Multi-\ac{GPU} configurations, allowing the rendering load to be split by dedicating a separate \ac{GPU} to compute each eye's viewport independently. 

A critical challenge in executing native \ac{VR} rendering via the Python \ac{API} is the geometric mismatch between the display hardware and the simulation engine. Modern \ac{VR} hardware, such as the Meta Quest 3, uses an asymmetric \ac{FOV} \cite{wuDontBlockGround2021} to account for human facial anatomy, typically rendering a wider angle towards the outside and a narrower angle towards the nose. Conversely, \ac{CARLA} cameras strictly accept a single, symmetric \ac{FOV} parameter. Naively projecting \ac{CARLA}'s symmetric output onto asymmetric lenses results in excessive image overlap and severe stereoscopic double vision.

To resolve this without altering \ac{CARLA}'s underlying C++ source code, WalkoCARLA implements a mathematical geometric correction pipeline. The system queries the \ac{HMD}'s asymmetric angles ($\theta_{\text{left}}, \theta_{\text{right}}, \theta_{\text{top}}, \theta_{\text{bottom}}$) via OpenXR and commands \ac{CARLA} to render an expanded, symmetric $\text{FOV}_{\text{sym}}$ that fully encapsulates the largest required viewing angle:

\begin{equation} \label{eq:fovsym}
	\text{FOV}_{\text{sym}} = 2 \cdot \max(|\theta_{\text{left}}|, |\theta_{\text{right}}|)
\end{equation}

The image data generated on the nasal side is subsequently cropped dynamically during the OpenGL rendering loop, matching the exact asymmetric parameters expected by the headset. Crucially, this OpenGL-based dynamic cropping introduces zero CPU or rendering overhead, as the coordinate clipping is executed entirely within the graphics hardware pipeline. 

Beyond rendering geometry, the \ac{VR} Process must actively resolve a fundamental coordinate system mismatch between OpenXR and Unreal Engine to track head movements correctly. OpenXR uses a right-handed coordinate system ($-Z$ forward, $+Y$ up, $+X$ right), whereas \ac{CARLA} expects left-handed coordinates ($+X$ forward, $+Z$ up, $+Y$ right). To synchronize the \ac{HMD} pose to the virtual cameras without inducing gimbal lock or disorientation, WalkoCARLA transforms the raw OpenXR tracking quaternions into intrinsic \texttt{YZX} Euler angles, negating the pitch and roll axes to invert the handedness before applying them via the \ac{CARLA} \ac{API}.

To further reduce rendering latency, the pipeline optimizes texture uploads to the \ac{GPU}. By using OpenGL \acp{PBO} and delegating the heavy \texttt{BGRA} to \texttt{RGB} format conversion directly to the \ac{GPU} instead of processing it via NumPy arrays, the architecture significantly reduces frame processing time, elevating the rendering loop from seven to 36\,\ac{FPS} on an average desktop computer.

\subsubsection{Pose Process}

The most critical element for naturalistic \ac{HITL} interaction is the Pose Process, which features a fully implemented 6-\ac{DOF} estimation pipeline. Because this pipeline relies on vision-based tracking algorithms rather than expensive, specialized full-body tracking suits, it remains accessible and can support older or standard consumer-grade hardware. 

\begin{figure*}[tbp]
	\centering
	\includegraphics[width=\textwidth]{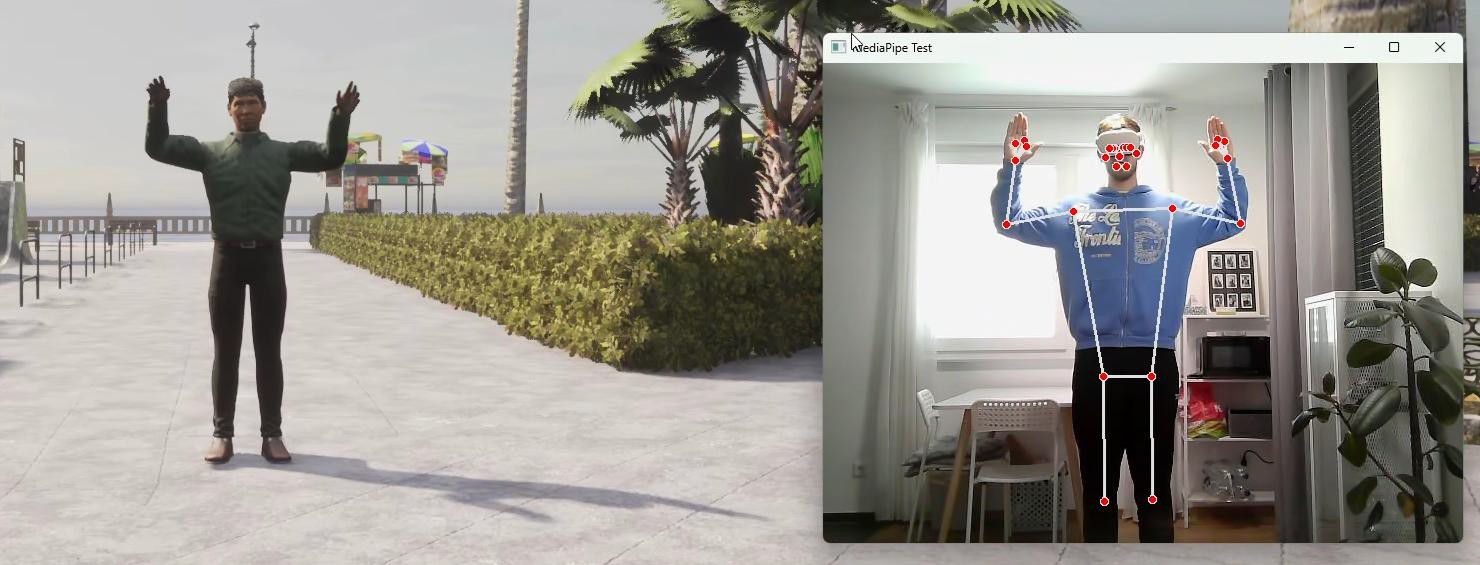} 
	\caption{Real-time pose estimation mapping. Three-dimensional landmarks are extracted from a standard RGB webcam via MediaPipe (right) and kinematically transformed to manipulate the skeletal mesh of the pedestrian avatar in the \ac{CARLA} simulation (left).}
	\label{fig:pose_estimation}
\end{figure*}

To bridge the behavioral gap between artificial \acp{NPC} and authentic human body language, WalkoCARLA integrates Google's MediaPipe framework \cite{lugaresiMediaPipeFrameworkBuilding2019}. Using a standard webcam, the system extracts 33 three-dimensional skeletal landmarks of the user in real time, as illustrated in Figure~\ref{fig:pose_estimation}. Because these raw coordinates are bound to the absolute, global space of the physical camera, they must be mathematically transformed before they can animate the \ac{CARLA} avatar. 

The system constructs a local, relative coordinate system for the user's torso. It calculates a shoulder vector $v_\text{shoulder}$ and a body longitudinal vector $v_\text{body}$, extending from the neck to the mid-hip landmark, taking the cross product of the two to determine the depth normal vector of the body $v_\text{back}$. 

\begin{equation}
	v_\text{back}=v_\text{shoulder}\times v_\text{body}
\end{equation}

Using these vectors, an optimal rotation matrix $R_\text{body}$ is derived, describing the exact orientation of the torso. By applying the inverse of this matrix ($R_\text{body}^{-1}$) to the vectors of the arms, the arm movements are systematically isolated into the local coordinate space of the body. 

\begin{equation}
	v_\text{arm,local}=R_\text{body}^{-1}\cdot v_\text{arm,global}
\end{equation}

These relative directional vectors are then converted into Euler angles (Yaw, Pitch, Roll) for spherical joints like the shoulder, and 1D hinge angles for the elbows. These kinematic parameters are continuously streamed via the \ac{API} to update the skeletal mesh of the ego-pedestrian. 

To prevent the computationally demanding vision inference from bottlenecking the critical \ac{VR} rendering loop, the Pose Process is executed entirely in parallel via Python multiprocessing. This bi-directional avatar synchronization ensures that drivers in DrivoCARLA can accurately perceive where the pedestrian is looking and gesturing. Capturing this non-verbal communication is a critical factor for studying naturalistic traffic negotiations that \ac{AI} models fundamentally fail to replicate.

\section{Conclusion and Future Work}
\label{sec:con}

Developing automated driving systems for urban environments requires detailed insight into mixed-traffic dynamics. \ac{HITL} simulators provide a safe and repeatable method for investigating such interactions without exposing participants to physical risk. However, a key limitation of traditional simulator setups has often been their operational isolation; studying mixed traffic benefits fundamentally from the synchronous interaction of multiple human actors, including \acp{VRU}.

This paper introduced the CARLAverse, a highly modular, distributed, and multimodal simulation ecosystem. Building upon previous work that established a hardware-agnostic abstraction layer for standalone driving simulators \cite{reblingHighlyModularImmersiveHumanintheLoop2025}, the CARLAverse expands this concept into a networked, multi-agent simulation ecosystem. By integrating vehicle simulators (DrivoCARLA), bicycle simulators (CycloCARLA), and pedestrian virtual reality nodes (WalkoCARLA) into a shared, centrally coordinated virtual world, the framework addresses the behavioral gap typically exhibited by \ac{AI}-driven \acp{NPC}.

A core technical contribution of this work is the implementation of a distributed physics architecture. By strictly decoupling the computationally intensive global traffic simulation (handled by a central server) from the latency-critical ego dynamics (calculated locally on dedicated local nodes), the CARLAverse is designed to reduce latency. Furthermore, by transitioning from middleware-routed control signals to direct \ac{CARLA} \ac{API} communication, the system supports high-frequency local control loops necessary for realistic haptic feedback and motion cueing, while using \ac{ROS} exclusively as a background data-logging engine. Enhanced by cross-cutting technologies such as rapid \ac{YAML}-based reconfiguration and the spatial CARLAcoustics engine, the CARLAverse is designed to provide an immersive, scalable, and reproducible platform for advanced traffic research.

\subsection{Scientific Utility}

The primary scientific use of the CARLAverse lies in its capacity to facilitate ecologically valid, multi-agent behavioral research that transcends the physical boundaries of a single laboratory. By providing a scalable, low-latency framework for synchronized interactions between diverse human actors, the ecosystem enables researchers to conduct naturalistic studies on critical factors such as social negotiations, intent recognition, and trust dynamics in mixed traffic scenarios. This environment addresses the behavioral gap inherent in purely \ac{AI}-driven \acp{NPC} and supports the collection of human response data in controlled mixed-traffic interactions. Furthermore, its hardware-agnostic, open-source architecture lowers the financial and technical barriers to entry, thereby enabling cross-institutional collaboration and fostering rigorous reproducibility in complex \ac{HITL} experiments.

\subsection{Future Work}

While the foundational architecture of the CARLAverse has been successfully implemented and deployed, several avenues for future research, validation, and system enhancement remain.

First, the physical models underlying the \ac{VRU} nodes require empirical validation. The longitudinal bicycle dynamics and force feedback implemented in CycloCARLA have been mathematically validated in a dedicated study \cite{reblingCycloCARLADesignImplementation2026} (to appear). Similarly, the spatial audio propagation and frequency masking algorithms within the CARLAcoustics module will be evaluated in an upcoming publication to quantify their impact on participant immersion and localization accuracy.

To further elevate the physical immersion of the simulator nodes, several hardware extensions are currently under development. For the bicycle simulator, a motion platform is being built to simulate uneven surfaces and slopes, alongside a dedicated wind simulation system. Concurrently, an omnidirectional treadmill is being integrated to support natural, unconstrained locomotion for the pedestrian node.

Enhancing visual fidelity and performance is another critical focus area. A rigorous evaluation of the current \ac{VR} render pipeline is planned, which will systematically compare it against different rendering methodologies. As part of this effort, the potential of using neural rendering techniques as an alternative to the default \ac{CARLA} render engine is being actively investigated to achieve higher levels of photorealism.

With the infrastructure now operational and continuously evolving, future work will also focus on conducting large-scale, synchronous \ac{HITL} experiments. Extending previous single-node evaluations \cite{reblingMindGapQuantifying2026}, planned studies will use standardized OpenSCENARIO files \cite{associationforstandardizationofautomation&measuringsystemsASAMOpenSCENARIOXML2024} to investigate yielding behaviors and trust dynamics in three-way negotiations between human-driven vehicles, human cyclists, and pedestrians at urban intersections. Furthermore, future work will explore the feasibility of fine-tuning \ac{FM}-based Sim-Agents using these data.

Finally, in alignment with the principles of open science, while core modules and tooling are already publicly available, the ongoing effort focuses on completing full system integration and releasing all remaining components of the CARLAverse framework. By providing the complete suite of hardware-agnostic Python modules, \ac{YAML} templates, and synchronization scripts to the global research community, we aim to lower the barrier to entry for immersive, connected simulator research and foster collaborative advancements in the field of automated mobility

\section*{Acknowledgments}

This work was developed in the projects KIIWI (reference number: 16DHBKI060) which is funded by the German Federal Ministry of Research, Technology and Space (BMFTR) and Real2Sim (reference number: BW7\_2199/2) which is funded by the Baden-Wuerttemberg Ministry of Economic Affairs, Labor and Tourism.

During the preparation of this work, the authors used Google Gemini for language editing and to assist in generating visual assets (e.g., project logos). The authors reviewed and edited the content and take full responsibility for the publication.

\section*{Definitions, Acronyms, Abbreviations}
\vspace{7pt}
\begin{small}
\begin{quote}
\begin{acronym}[AUTOSAR\quad\quad]
	\acro{ADAS}{Advanced Driver Assistance System}
	\acroplural{ADAS}{Advanced Driver Assistance Systems}
	\acro{AI}{Artificial Intelligence}
	\acro{API}{Application Programming Interface}
	\acro{AUTOSAR}{AUTomotive Open System ARchitecture}
	\acro{AV}{Autonomous Vehicle}
	\acro{BLE}{Bluetooth Low Energy}
	\acro{CAN}{Controller Area Network}
	\acro{CARLA}{Car-Learning-to-Act}
	\acro{CARMEn}{CARla-Based, Multi-Agent Immersive Road Environment Simulation}
	\acro{CAVE}{Cave Automatic Virtual Environment}
	\acro{CC}{Cruise Control}
	\acro{CDA-HDS}{Cooperative Driving Automation - Highway Driving Simulator}
	\acro{DC}{Direct Current}
	\acro{DDS}{Data Distribution Service}
	\acro{DMAVA}{Distributed Multi-Autonomous Vehicle Architecture}
	\acro{DOF}{Degree-of-Freedom}
	\acro{FFB}{Force Feedback}
	\acro{FM}{Foundation Model}
	\acro{FOV}{Field of View}
	\acro{FPS}{Frames per Second}
	\acro{GPU}{Graphics Processing Unit}
	\acro{GUI}{Graphical User Interface}
	\acro{HAL}{Hardware Abstraction Layer}
	\acro{HITL}{Human-in-the-Loop}
	\acro{HMD}{Head-Mounted Display}
	\acro{HMI}{Human-Machine Interface}
	\acro{HTML}{HyperText Markup Language}
	\acro{HTTP}{Hypertext Transfer Protocol}
	\acro{IMU}{inertial measurement unity}
	\acro{IP}{Internet Protocol}
	\acro{IPC}{Inter-process Communication}
	\acro{LCA}{Lane Change Assist}
	\acro{LGSVL}{LG Silicon Valley Lab}
	\acro{LKAS}{Lane Keeping Assist System}
	\acro{LPF}{low-pass filter}
	\acro{LUT}{Lookup Table}
	\acro{MCA}{Motion Cueing Algorithm}
	\acro{NPC}{Non-Player Character}
	\acro{PBO}{Pixel Buffer Object}
	\acro{ROS}{Robot Operating System}
	\acro{RPC}{Remote Procedure Call}
	\acro{RPM}{Revolutions per Minute}
	\acro{SDL}{Simple DirectMedia Layer}
	\acro{SIL}{Software-in-the-Loop}
	\acro{SUMO}{Simulation of Urban MObility}
	\acro{TCP}{Transmission Control Protocol}
	\acro{TENA}{Test and Training Enabling Architecture}
	\acro{TraCI}{Traffic Control Interface}
	\acro{UDP}{User Datagram Protocol}
	\acro{UE}{Unreal Engine}
	\acro{UI}{User Interface}
	\acro{USB}{Universal Serial Bus}
	\acro{V2X}{Vehicle-to-Everything}
	\acro{VR}{Virtual Reality}
	\acro{VRU}{Vulnerable Road User}
	\acro{WAN}{Wide Area Network}
	\acro{XR}{Extended Reality}
	\acro{YAML}{YAML Ain't Markup Language}
\end{acronym}
\end{quote}
\end{small}

\newpage

\printbibliography

@standard{associationforstandardizationofautomation&measuringsystemsASAMOpenSCENARIOXML2024,
  title = {{{ASAM OpenSCENARIO XML}}},
  author = {{Association for Standardization of Automation \& Measuring Systems}},
  date = {2024},
  url = {https://www.asam.net/standards/detail/openscenario/},
  urldate = {2026-09-10},
  version = {1.3.1}
}

@article{azfarTrafficCosimulationFramework2025,
  title = {Traffic Co-Simulation Framework Empowered by Infrastructure Camera Sensing and Reinforcement Learning},
  author = {Azfar, Talha and Huang, Kaicong and Tracy, Andrew and Misiewicz, Sandra and Liu, Chenxi and Ke, Ruimin},
  date = {2025-09-15},
  journaltitle = {Journal of Intelligent Transportation Systems},
  shortjournal = {J. Intell. Transp. Syst.},
  pages = {1--19},
  issn = {1547-2450, 1547-2442},
  doi = {10.1080/15472450.2025.2559410},
  url = {https://www.tandfonline.com/doi/full/10.1080/15472450.2025.2559410},
  urldate = {2026-09-10},
  langid = {english}
}

@online{azimiSimulationAllStepbyStep2025,
  title = {Simulation for {{All}}: {{A Step-by-Step Cookbook}} for {{Developing Human-Centered Multi-Agent Transportation Simulators}}},
  shorttitle = {Simulation for {{All}}},
  author = {Azimi, Shiva and Tavakoli, Arash},
  date = {2025},
  doi = {10.48550/ARXIV.2507.09367},
  url = {https://arxiv.org/abs/2507.09367},
  urldate = {2026-09-10},
  version = {2}
}

@article{beckersJOANFrameworkHumanautomated2023,
  title = {{{JOAN}}: A Framework for Human-Automated Vehicleinteraction Experiments in a Virtual Reality Driving Simulator},
  shorttitle = {{{JOAN}}},
  author = {Beckers, Niek and Siebinga, Olger and Giltay, Joris and Van Der Kraan, André},
  date = {2023-02-01},
  journaltitle = {Journal of Open Source Software},
  shortjournal = {J. Open Source Softw.},
  volume = {8},
  number = {82},
  pages = {4250},
  issn = {2475-9066},
  doi = {10.21105/joss.04250},
  url = {https://joss.theoj.org/papers/10.21105/joss.04250},
  urldate = {2026-09-10}
}

@article{bimbrawAutonomousCarsPresent2015,
  title = {Autonomous {{Cars}}: {{Past}}, {{Present}} and {{Future}} - {{A Review}} of the {{Developments}} in the {{Last Century}}, the {{Present Scenario}} and the {{Expected Future}} of {{Autonomous Vehicle Technology}}},
  shorttitle = {Autonomous {{Cars}}},
  author = {Bimbraw, Keshav},
  date = {2015-01-01},
  journaltitle = {ICINCO 2015 - 12th International Conference on Informatics in Control, Automation and Robotics, Proceedings},
  shortjournal = {ICINCO 2015 - 12th Int. Conf. Inform. Control Autom. Robot. Proc.},
  volume = {1},
  pages = {191--198},
  doi = {10.5220/0005540501910198}
}

@article{cipelliDriverintheloopSimulationsParametric2008,
  title = {Driver-in-the-Loop Simulations with Parametric Car Models},
  author = {Cipelli, Marco and Schiehlen, Werner and Cheli, Federico},
  date = {2008-09-01},
  journaltitle = {Vehicle System Dynamics},
  shortjournal = {Veh. Syst. Dyn.},
  volume = {46},
  pages = {33--48},
  issn = {0042-3114},
  doi = {10.1080/00423110701882280},
  issue = {sup1}
}

@article{creweSLAVSimFrameworkSelfLearning2023,
  title = {{{SLAV-Sim}}: {{A Framework}} for {{Self-Learning Autonomous Vehicle Simulation}}},
  shorttitle = {{{SLAV-Sim}}},
  author = {Crewe, Jacob and Humnabadkar, Aditya and Liu, Yonghuai and Ahmed, Amr and Behera, Ardhendu},
  date = {2023-10-23},
  journaltitle = {Sensors},
  volume = {23},
  number = {20},
  pages = {8649},
  issn = {1424-8220},
  doi = {10.3390/s23208649},
  url = {https://www.mdpi.com/1424-8220/23/20/8649},
  urldate = {2026-09-10},
  langid = {english}
}

@inproceedings{crosatoVirtualRealityFramework2024,
  title = {A {{Virtual Reality Framework}} for {{Human-Driver Interaction Research}}: {{Safe}} and {{Cost-Effective Data Collection}}},
  shorttitle = {A {{Virtual Reality Framework}} for {{Human-Driver Interaction Research}}},
  booktitle = {Proceedings of the 2024 {{ACM}}/{{IEEE International Conference}} on {{Human-Robot Interaction}}},
  author = {Crosato, Luca and Wei, Chongfeng and Ho, Edmond S. L. and Shum, Hubert P. H. and Sun, Yuzhu},
  date = {2024-03-11},
  pages = {167--174},
  publisher = {ACM},
  location = {Boulder CO USA},
  doi = {10.1145/3610977.3634923},
  url = {https://dl.acm.org/doi/10.1145/3610977.3634923},
  urldate = {2026-09-10},
  eventtitle = {{{HRI}} '24: {{ACM}}/{{IEEE International Conference}} on {{Human-Robot Interaction}}},
  isbn = {979-8-4007-0322-5},
  langid = {english}
}

@article{defilippoModularArchitectureDriving2014,
  title = {A Modular Architecture for a Driving Simulator Based on the {{FDMU}} Approach},
  author = {De Filippo, F. and Stork, A. and Schmedt, H. and Bruno, F.},
  date = {2014-05},
  journaltitle = {International Journal on Interactive Design and Manufacturing (IJIDeM)},
  shortjournal = {Int. J. Interact. Des. Manuf. IJIDeM},
  volume = {8},
  number = {2},
  pages = {139--150},
  issn = {1955-2513, 1955-2505},
  doi = {10.1007/s12008-013-0182-3},
  langid = {english}
}

@inproceedings{deyGrokWalksPortableVirtual2024,
  title = {{{GrokWalks}}: {{A Portable Virtual Reality Platform}} to {{Facilitate Studying Driver-Pedestrian Interactions}}},
  shorttitle = {{{GrokWalks}}},
  booktitle = {Adjunct {{Proceedings}} of the 16th {{International Conference}} on {{Automotive User Interfaces}} and {{Interactive Vehicular Applications}}},
  author = {Dey, Debargha and Goedicke, David and Yang, Chishang and Sirkin, David and Currano, Rebecca and Ju, Wendy},
  date = {2024-09-22},
  pages = {284--288},
  publisher = {ACM},
  location = {Stanford CA USA},
  doi = {10.1145/3641308.3685013},
  url = {https://dl.acm.org/doi/10.1145/3641308.3685013},
  urldate = {2026-09-10},
  eventtitle = {{{AutomotiveUI}} '24: 16th {{International Conference}} on {{Automotive User Interfaces}} and {{Interactive Vehicular Applications}}},
  isbn = {979-8-4007-0520-5},
  langid = {english}
}

@inproceedings{fischerModularScalableDriving2014,
  title = {Modular and Scalable Driving Simulator Hardware and Software for the Development of Future Driver Assistence and Automation Systems},
  booktitle = {New Developments in Driving Simulation Design and Experiments},
  author = {Fischer, Martin and Richter, Andreas and Schindler, Julian and Plättner, Jens and Temme, Gerald and Kelsch, Johann and Assmann, Dirk and Köster, Frank},
  editor = {Kemeny, Andras and Espié, Stéphane and Mérienne, Frédéric},
  date = {2014-09-04},
  pages = {223--229},
  location = {Paris, Frankreich},
  issn = {0769-0266},
  url = {https://elib.dlr.de/90638/},
  urldate = {2026-09-10},
  eventtitle = {Driving Simulation Conference 2014},
  langid = {ngerman}
}

@report{gillespieTrustArtificialIntelligence2023,
  title = {Trust in {{Artificial Intelligence}}: {{A}} Global Study},
  shorttitle = {Trust in {{Artificial Intelligence}}},
  author = {Gillespie, Nicole and Lockey, Steven and Curtis, Caitlin and Pool, Javad and {Ali Akbari}},
  date = {2023-02-16},
  institution = {The University of Queensland; KPMG Australia},
  location = {Brisbane, Australia},
  url = {http://doi.org/10.14264/00d3c94},
  urldate = {2026-09-10},
  langid = {english}
}

@thesis{hanParallelDevelopmentTesting2024,
  title = {A {{Parallel Development}} and {{Testing Framework}} for {{Cooperative Driving Automation}}},
  author = {Han, Xu},
  date = {2024},
  institution = {UCLA},
  url = {https://escholarship.org/uc/item/0bd7d5wv},
  urldate = {2026-09-10},
  langid = {english}
}

@article{heStableHapticRendering2013,
  title = {Stable {{Haptic Rendering For Physics Engines Using Inter-Process Communication}} and {{Remote Virtual Coupling}}},
  author = {He, Xue-Jian and Choi, Kup-Sze},
  date = {2013},
  journaltitle = {International Journal of Advanced Computer Science and Applications},
  shortjournal = {Int. J. Adv. Comput. Sci. Appl.},
  volume = {4},
  number = {1},
  issn = {2158107X, 21565570},
  doi = {10.14569/IJACSA.2013.040103},
  url = {http://thesai.org/Publications/ViewPaper?Volume=4&Issue=1&Code=IJACSA&SerialNo=3},
  urldate = {2026-09-10},
  langid = {english}
}

@online{huangSkyDriveDistributedMultiAgent2025,
  title = {Sky-{{Drive}}: {{A Distributed Multi-Agent Simulation Platform}} for {{Human-AI Collaborative}} and {{Socially-Aware Future Transportation}}},
  shorttitle = {Sky-{{Drive}}},
  author = {Huang, Zilin and Sheng, Zihao and Wan, Zhengyang and Qu, Yansong and Luo, Yuhao and Wang, Boyue and Li, Pei and Chen, Yen-Jung and Chen, Jiancong and Long, Keke and Meng, Jiayi and Leng, Yue and Chen, Sikai},
  date = {2025},
  doi = {10.48550/ARXIV.2504.18010},
  url = {https://arxiv.org/abs/2504.18010},
  urldate = {2026-09-10},
  version = {2}
}

@incollection{ihemedu-steinkeDevelopmentEvaluationVirtual2015,
  title = {Development and {{Evaluation}} of a {{Virtual Reality Driving Simulator}}},
  booktitle = {Mensch Und {{Computer}} 2015 - {{Workshopband}}},
  author = {Ihemedu-Steinke, Quinate Chioma and Sirim, Demet and Erbach, Rainer and Halady, Prashanth and Meixner, Gerrit},
  editor = {Weisbecker, Anette and Burmester, Michael and Schmidt, Albrecht},
  date = {2015-08-28},
  pages = {491--500},
  publisher = {DE GRUYTER},
  doi = {10.1515/9783110443905-070},
  url = {https://www.degruyter.com/document/doi/10.1515/9783110443905-070/html},
  urldate = {2026-09-10},
  isbn = {978-3-11-044333-2}
}

@online{islamDMAVADistributedMultiAutonomous2026,
  title = {{{DMAVA}}: {{Distributed Multi-Autonomous Vehicle Architecture Using Autoware}}},
  shorttitle = {{{DMAVA}}},
  author = {Islam, Zubair and El-Darieby, Mohamed},
  date = {2026},
  doi = {10.48550/ARXIV.2601.16336},
  url = {https://arxiv.org/abs/2601.16336},
  urldate = {2026-09-10},
  version = {2}
}

@article{kellyWhatFactorsContribute2023,
  title = {What Factors Contribute to the Acceptance of Artificial Intelligence? {{A}} Systematic Review},
  shorttitle = {What Factors Contribute to the Acceptance of Artificial Intelligence?},
  author = {Kelly, Sage and Kaye, Sherrie-Anne and Oviedo-Trespalacios, Oscar},
  date = {2023-02},
  journaltitle = {Telematics and Informatics},
  shortjournal = {Telemat. Inform.},
  volume = {77},
  pages = {101925},
  issn = {07365853},
  doi = {10.1016/j.tele.2022.101925},
  langid = {english}
}

@article{kemenyEvaluatingPerceptionDriving2003,
  title = {Evaluating Perception in Driving Simulation Experiments},
  author = {Kemeny, Andras and Panerai, Francesco},
  date = {2003-01-01},
  journaltitle = {Trends in Cognitive Sciences},
  shortjournal = {Trends Cogn. Sci.},
  volume = {7},
  number = {1},
  pages = {31--37},
  issn = {1364-6613},
  doi = {10.1016/S1364-6613(02)00011-6},
  url = {https://www.sciencedirect.com/science/article/pii/S1364661302000116},
  urldate = {2026-09-10}
}

@report{khronos_openxr,
  title = {{{OpenXR}} Specification},
  author = {{The Khronos Group}},
  date = {2024},
  institution = {The Khronos Group Inc.},
  url = {https://www.khronos.org/openxr/},
  urldate = {2026-09-10}
}

@online{kronauerLatencyAnalysisROS22021,
  title = {Latency {{Analysis}} of {{ROS2 Multi-Node Systems}}},
  author = {Kronauer, Tobias and Pohlmann, Joshwa and Matthe, Maximilian and Smejkal, Till and Fettweis, Gerhard},
  date = {2021},
  doi = {10.48550/ARXIV.2101.02074},
  url = {https://arxiv.org/abs/2101.02074},
  urldate = {2026-09-10},
  version = {3}
}

@inproceedings{kruegerSILABTaskOriented2005,
  title = {{{SILAB}} - {{A Task Oriented Driving Simulation}}},
  author = {Krueger, H. and Grein, Martina and Kaußner, Armin and Mark, Christian},
  date = {2005},
  url = {https://api.semanticscholar.org/CorpusID:11019771},
  urldate = {2026-09-10}
}

@incollection{lehsingUrbanInteractionGetting2018,
  title = {Urban {{Interaction}} – {{Getting Vulnerable Road Users}} into {{Driving Simulation}}},
  booktitle = {{{UR}}:{{BAN Human Factors}} in {{Traffic}}},
  author = {Lehsing, Christian and Feldstein, Ilja T.},
  editor = {Bengler, Klaus and Drüke, Julia and Hoffmann, Silja and Manstetten, Dietrich and Neukum, Alexandra},
  date = {2018},
  pages = {347--362},
  publisher = {Springer Fachmedien Wiesbaden},
  location = {Wiesbaden},
  doi = {10.1007/978-3-658-15418-9_19},
  url = {http://link.springer.com/10.1007/978-3-658-15418-9_19},
  urldate = {2026-09-10},
  isbn = {978-3-658-15417-2},
  langid = {english}
}

@inproceedings{lindnerCoupledDrivingSimulator2022,
  title = {A Coupled Driving Simulator to Investigate the Interaction between Bicycles and Automated Vehicles},
  booktitle = {2022 {{IEEE}} 25th {{International Conference}} on {{Intelligent Transportation Systems}} ({{ITSC}})},
  author = {Lindner, Johannes and Keler, Andreas and Grigoropoulos, Georgios and Malcolm, Patrick and Denk, Florian and Brunner, Pascal and Bogenberger, Klaus},
  date = {2022-10-08},
  pages = {1335--1341},
  publisher = {IEEE},
  location = {Macau, China},
  doi = {10.1109/ITSC55140.2022.9922400},
  url = {https://ieeexplore.ieee.org/document/9922400/},
  urldate = {2026-09-10},
  eventtitle = {2022 {{IEEE}} 25th {{International Conference}} on {{Intelligent Transportation Systems}} ({{ITSC}})},
  isbn = {978-1-6654-6880-0}
}

@inproceedings{liSurveyVirtualMachine2010,
  title = {A {{Survey}} of {{Virtual Machine System}}: {{Current Technology}} and {{Future Trends}}},
  booktitle = {2010 {{Third International Symposium}} on {{Electronic Commerce}} and {{Security}}},
  author = {Li, Yunfa and Li, Wanqing and Jiang, Congfeng},
  date = {2010},
  pages = {332--336},
  doi = {10.1109/ISECS.2010.80}
}

@inproceedings{lopezMicroscopicTrafficSimulation2018,
  title = {Microscopic {{Traffic Simulation}} Using {{SUMO}}},
  booktitle = {2018 21st {{International Conference}} on {{Intelligent Transportation Systems}} ({{ITSC}})},
  author = {Lopez, Pablo Alvarez and Wiessner, Evamarie and Behrisch, Michael and Bieker-Walz, Laura and Erdmann, Jakob and Flotterod, Yun-Pang and Hilbrich, Robert and Lucken, Leonhard and Rummel, Johannes and Wagner, Peter},
  date = {2018-11},
  pages = {2575--2582},
  publisher = {IEEE},
  location = {Maui, HI},
  doi = {10.1109/ITSC.2018.8569938},
  url = {https://ieeexplore.ieee.org/document/8569938/},
  urldate = {2026-09-10},
  eventtitle = {2018 21st {{International Conference}} on {{Intelligent Transportation Systems}} ({{ITSC}})},
  isbn = {978-1-7281-0321-1}
}

@online{lugaresiMediaPipeFrameworkBuilding2019,
  title = {{{MediaPipe}}: {{A Framework}} for {{Building Perception Pipelines}}},
  shorttitle = {{{MediaPipe}}},
  author = {Lugaresi, Camillo and Tang, Jiuqiang and Nash, Hadon and McClanahan, Chris and Uboweja, Esha and Hays, Michael and Zhang, Fan and Chang, Chuo-Ling and Yong, Ming Guang and Lee, Juhyun and Chang, Wan-Teh and Hua, Wei and Georg, Manfred and Grundmann, Matthias},
  date = {2019},
  doi = {10.48550/ARXIV.1906.08172},
  url = {https://arxiv.org/abs/1906.08172},
  urldate = {2026-09-10},
  version = {1}
}

@inproceedings{machadoCARMEnCARlaBasedMultiAgent2026,
  title = {{{CARMEn}}: {{CARla-Based}}, {{Multi-Agent Immersive Road Environment Simulation}}:},
  shorttitle = {{{CARMEn}}},
  booktitle = {Proceedings of the 21st {{International Conference}} on {{Computer Graphics}}, {{Interaction}} and {{Visualization Theory}} and {{Applications}}},
  author = {Machado, Dário and Pereira, Frederico and Monteiro, Sérgio and Louro, Luís and Almeida, Raul and Gomes, Mónica and Freitas, Elisabete and Bicho, Estela and Sousa, Emanuel},
  date = {2026},
  pages = {120--130},
  publisher = {{SCITEPRESS - Science and Technology Publications}},
  location = {Marbella, Spain},
  doi = {10.5220/0014467600004728},
  url = {https://www.scitepress.org/DigitalLibrary/Link.aspx?doi=10.5220/0014467600004728},
  urldate = {2026-09-10},
  eventtitle = {21st {{International Conference}} on {{Computer Graphics}}, {{Interaction}} and {{Visualization Theory}} and {{Applications}}},
  isbn = {978-989-758-803-7}
}

@incollection{massaHardwareAbstractionLayer2003,
  title = {The {{Hardware Abstraction Layer}}},
  booktitle = {Embedded Software Development with E-{{Cos}}},
  author = {Massa, Anthony J.},
  date = {2003},
  series = {Bruce {{Perens}}' {{Open}} Source Series},
  publisher = {Prentice Hall},
  location = {Upper Saddle River, NJ},
  isbn = {978-0-13-035473-0}
}

@article{molnarUnderstandingTrustAcceptance2018,
  title = {Understanding Trust and Acceptance of Automated Vehicles: {{An}} Exploratory Simulator Study of Transfer of Control between Automated and Manual Driving},
  shorttitle = {Understanding Trust and Acceptance of Automated Vehicles},
  author = {Molnar, Lisa J. and Ryan, Lindsay H. and Pradhan, Anuj K. and Eby, David W. and St. Louis, Renée M. and Zakrajsek, Jennifer S.},
  date = {2018-10},
  journaltitle = {Transportation Research Part F: Traffic Psychology and Behaviour},
  shortjournal = {Transp. Res. Part F Traffic Psychol. Behav.},
  volume = {58},
  pages = {319--328},
  issn = {13698478},
  doi = {10.1016/j.trf.2018.06.004},
  langid = {english}
}

@misc{muskInterviewElonMusk2016,
  title = {Interview with {{Elon Musk}} at {{Code Conference}} 2016},
  namea = {Musk, Elon and Swisher, Kara and Mossberg, Walt},
  nameatype = {collaborator},
  date = {2016},
  url = {https://www.youtube.com/watch?v=wsixsRI-Sz4},
  urldate = {2026-09-10},
  langid = {english}
}

@online{neisCARJANAgentBasedGeneration2025,
  title = {{{CARJAN}}: {{Agent-Based Generation}} and {{Simulation}} of {{Traffic Scenarios}} with {{AJAN}}},
  shorttitle = {{{CARJAN}}},
  author = {Neis, Leonard Frank and Antakli, Andre and Klusch, Matthias},
  date = {2025},
  doi = {10.48550/ARXIV.2508.21411},
  url = {https://arxiv.org/abs/2508.21411},
  urldate = {2026-09-10},
  version = {1}
}

@inproceedings{noseworthyTestTrainingEnabling2008a,
  title = {The {{Test}} and {{Training Enabling Architecture}} ({{TENA}}) {{Supporting}} the {{Decentralized Development}} of {{Distributed Applications}} and {{LVC Simulations}}},
  booktitle = {2008 12th {{IEEE}}/{{ACM International Symposium}} on {{Distributed Simulation}} and {{Real-Time Applications}}},
  author = {Noseworthy, J. Russell},
  date = {2008-10},
  pages = {259--268},
  publisher = {IEEE},
  location = {Vancouver, BC, Canada},
  doi = {10.1109/DS-RT.2008.35},
  url = {http://ieeexplore.ieee.org/document/4700128/},
  urldate = {2026-09-10},
  eventtitle = {2008 12th {{IEEE International Symposium}} on {{Distributed Simulation}} and {{Real-Time Applications}} ({{DS-RT}})},
  isbn = {978-0-7695-3425-1}
}

@article{padmajaExplorationIssuesChallenges2023,
  title = {Exploration of Issues, Challenges and Latest Developments in Autonomous Cars},
  author = {Padmaja, B. and family=Moorthy, given=Ch. V. K. N. S. N., given-i={{Ch}}VKNSN and Venkateswarulu, N. and Bala, Myneni Madhu},
  date = {2023-05-06},
  journaltitle = {Journal of Big Data},
  shortjournal = {J. Big Data},
  volume = {10},
  number = {1},
  pages = {61},
  issn = {2196-1115},
  doi = {10.1186/s40537-023-00701-y},
  langid = {english}
}

@article{perezARPEDFrameworkAugmented2019,
  title = {{{AR-PED}}: {{A}} Framework of Augmented Reality Enabled Pedestrian-in-the-Loop Simulation},
  shorttitle = {{{AR-PED}}},
  author = {Perez, Daniel and Hasan, Mahmud and Shen, Yuzhong and Yang, Hong},
  date = {2019-07},
  journaltitle = {Simulation Modelling Practice and Theory},
  shortjournal = {Simul. Model. Pract. Theory},
  volume = {94},
  pages = {237--249},
  issn = {1569190X},
  doi = {10.1016/j.simpat.2019.03.005},
  url = {https://linkinghub.elsevier.com/retrieve/pii/S1569190X19300292},
  urldate = {2026-09-10},
  langid = {english}
}

@misc{pygame,
  title = {Pygame},
  author = {Shinners, Pete and Lindstrom, Lenard and Dudfield, René and others},
  date = {2026},
  url = {https://www.pygame.org/},
  urldate = {2026-09-10},
  organization = {Pygame Community}
}

@incollection{quanteMeasuringDescribingCooperation2024,
  title = {Measuring and {{Describing Cooperation Between Road Users}}—{{Results}} from {{CoMove}}},
  booktitle = {Cooperatively {{Interacting Vehicles}}},
  author = {Quante, Laura and Stoll, Tanja and Baumann, Martin and Diera, Andor and Földes-Cappellotto, Noèmi and Jipp, Meike and Schießl, Caroline},
  editor = {Stiller, Christoph and Althoff, Matthias and Burger, Christoph and Deml, Barbara and Eckstein, Lutz and Flemisch, Frank},
  date = {2024},
  pages = {565--608},
  publisher = {Springer International Publishing},
  location = {Cham},
  doi = {10.1007/978-3-031-60494-2_20},
  url = {https://link.springer.com/10.1007/978-3-031-60494-2_20},
  urldate = {2026-09-10},
  isbn = {978-3-031-60493-5},
  langid = {english}
}

@inproceedings{quigleyROSOpensourceRobot2009,
  title = {{{ROS}}: An Open-Source {{Robot Operating System}}},
  booktitle = {{{IEEE International Conference}} on {{Robotics}} and {{Automation}}},
  author = {Quigley, Morgan and Conley, Ken and Gerkey, Brian and Faust, Josh and Foote, Tully and Leibs, Jeremy and Wheeler, Rob and Ng, Andrew},
  date = {2009-01},
  volume = {3},
  url = {https://api.semanticscholar.org/CorpusID:6324125},
  urldate = {2026-09-10}
}

@article{ramlallDevelopmentNetworkedMultiParticipant2025,
  title = {Development of a {{Networked Multi-Participant Driving Simulator}} with {{Synchronized EEG}} and {{Telemetry}} for {{Traffic Research}}},
  author = {Ramlall, Poorendra and Jones, Ethan and Roy, Subhradeep},
  date = {2025-07-10},
  journaltitle = {Systems},
  volume = {13},
  number = {7},
  pages = {564},
  issn = {2079-8954},
  doi = {10.3390/systems13070564},
  url = {https://www.mdpi.com/2079-8954/13/7/564},
  urldate = {2026-09-10},
  langid = {english}
}

@article{rampfModellingAutonomousVehicle2023,
  title = {Modelling Autonomous Vehicle Interactions with Bicycles in Traffic Simulation},
  author = {Rampf, Felix and Grigoropoulos, Georgios and Malcolm, Patrick and Keler, Andreas and Bogenberger, Klaus},
  date = {2023-01-04},
  journaltitle = {Frontiers in Future Transportation},
  shortjournal = {Front. Future Transp.},
  volume = {3},
  pages = {894148},
  issn = {2673-5210},
  doi = {10.3389/ffutr.2022.894148},
  url = {https://www.frontiersin.org/articles/10.3389/ffutr.2022.894148/full},
  urldate = {2026-09-10}
}

@inproceedings{rauchApproachHolisticModelling2023,
  title = {Approach to a {{Holistic Modelling}} of {{Cycling Dynamics}}},
  booktitle = {{{SIMUL2023}} - {{The Fifteenth International Conference}} on {{Advances}} in {{System}}},
  author = {Rauch, Yannick and Rall, Julia and Ruhe, Maximilian and Kriesten, Reiner},
  date = {2023-11-13},
  eventtitle = {{{SIMUL2023}} - {{The Fifteenth International Conference}} on {{Advances}} in {{System}}}
}

@inproceedings{reblingCycloCARLADesignImplementation2026,
  title = {{{CycloCARLA}}: {{Design}}, {{Implementation}}, and {{Evaluation}} of a {{High-Fidelity Bicycle Simulator}}},
  booktitle = {Proceedings of the Driving Simulation Conference 2026 Europe {{XR}}},
  author = {Rebling, Patrick and Rauch, Yannick and Sander, Nico and Nenninger, Philipp and Kriesten, Reiner},
  date = {2026},
  publisher = {Driving Simulation Association},
  eventtitle = {Accepted for Publication}
}

@inproceedings{reblingHighlyModularImmersiveHumanintheLoop2025,
  title = {Highly-{{Modular}} and {{Immersive Human-in-the-Loop Driving Simulators Using}} the {{CARLA Simulation Environment}}},
  author = {Rebling, Patrick and Beeh, Lars and Nenninger, Philipp and Kriesten, Reiner},
  date = {2025-09-28},
  pages = {9--14},
  url = {https://personales.upv.es/thinkmind/SIMUL/SIMUL_2025/simul_2025_1_20_50017.html},
  urldate = {2026-09-10},
  eventtitle = {{{SIMUL}} 2025, {{The Seventeenth International Conference}} on {{Advances}} in {{System Modeling}} and {{Simulation}}},
  isbn = {978-1-68558-300-2}
}

@inproceedings{reblingMindGapQuantifying2026,
  title = {Mind the {{Gap}}: {{Quantifying Behavioral Fidelity}} in {{CARLA Using Naturalistic Drone Data}}},
  shorttitle = {Mind the {{Gap}}},
  author = {Rebling, Patrick and Alphan, Metehan and Nenninger, Philipp},
  date = {2026-07-01},
  pages = {2026-01-0771},
  location = {Stuttgart, Germany},
  doi = {10.4271/2026-01-0771},
  url = {https://saemobilus.sae.org/papers/mind-gap-quantifying-behavioral-fidelity-carla-using-naturalistic-drone-data-2026-01-0771},
  urldate = {2026-07-10},
  eventtitle = {2026 {{Stuttgart International Symposium}}},
  langid = {english}
}

@article{rodsethNovelLowcostSolution2017,
  title = {A Novel Low-Cost Solution for Driving Assessment in Individuals with and without Disabilities},
  author = {Rodseth, Jakob and Washabaugh, Edward P. and Al Haddad, Ali and Kartje, Paula and Tate, Denise G. and Krishnan, Chandramouli},
  date = {2017-11},
  journaltitle = {Applied Ergonomics},
  shortjournal = {Appl. Ergon.},
  volume = {65},
  pages = {335--344},
  issn = {00036870},
  doi = {10.1016/j.apergo.2017.07.002},
  url = {https://linkinghub.elsevier.com/retrieve/pii/S0003687017301564},
  urldate = {2026-09-10},
  langid = {english}
}

@inproceedings{rongLGSVLSimulatorHigh2020,
  title = {{{LGSVL Simulator}}: {{A High Fidelity Simulator}} for {{Autonomous Driving}}},
  shorttitle = {{{LGSVL Simulator}}},
  booktitle = {2020 {{IEEE}} 23rd {{International Conference}} on {{Intelligent Transportation Systems}} ({{ITSC}})},
  author = {Rong, Guodong and Shin, Byung Hyun and Tabatabaee, Hadi and Lu, Qiang and Lemke, Steve and Mozeiko, Martins and Boise, Eric and Uhm, Geehoon and Gerow, Mark and Mehta, Shalin and Agafonov, Eugene and Kim, Tae Hyung and Sterner, Eric and Ushiroda, Keunhae and Reyes, Michael and Zelenkovsky, Dmitry and Kim, Seonman},
  date = {2020-09-20},
  pages = {1--6},
  publisher = {IEEE},
  location = {Rhodes, Greece},
  doi = {10.1109/ITSC45102.2020.9294422},
  url = {https://ieeexplore.ieee.org/document/9294422/},
  urldate = {2026-09-10},
  eventtitle = {2020 {{IEEE}} 23rd {{International Conference}} on {{Intelligent Transportation Systems}} ({{ITSC}})},
  isbn = {978-1-7281-4149-7}
}

@inproceedings{sabetiMADIVEMultiAgentDistributed2024,
  title = {{{MAD-IVE}}: {{Multi-Agent Distributed Immersive Virtual Environments}} for {{Vulnerable Road User Research}}—{{Potential}}, {{Challenges}}, and {{Requirements}}},
  shorttitle = {{{MAD-IVE}}},
  booktitle = {Computing in {{Civil Engineering}} 2023},
  author = {Sabeti, Sepehr and Tavakoli, Arash and Heydarian, Arsalan and Shoghli, Omidreza},
  date = {2024-01-25},
  pages = {1113--1120},
  publisher = {American Society of Civil Engineers},
  location = {Corvallis, Oregon},
  doi = {10.1061/9780784485248.133},
  url = {https://ascelibrary.org/doi/10.1061/9780784485248.133},
  urldate = {2026-09-10},
  eventtitle = {{{ASCE International Conference}} on {{Computing}} in {{Civil Engineering}} 2023},
  isbn = {978-0-7844-8524-8},
  langid = {english}
}

@article{sahaiCrossingStreetFront2022,
  title = {Crossing the Street in Front of an Autonomous Vehicle: {{An}} Investigation of Eye Contact between Drivengers and Vulnerable Road Users},
  shorttitle = {Crossing the Street in Front of an Autonomous Vehicle},
  author = {Sahaï, Aïsha and Labeye, Elodie and Caroux, Loïc and Lemercier, Céline},
  date = {2022-10-28},
  journaltitle = {Frontiers in Psychology},
  shortjournal = {Front. Psychol.},
  volume = {13},
  pages = {981666},
  issn = {1664-1078},
  doi = {10.3389/fpsyg.2022.981666},
  url = {https://www.frontiersin.org/articles/10.3389/fpsyg.2022.981666/full},
  urldate = {2026-09-10}
}

@article{saidiTransportDelayCharacterization2010,
  title = {Transport {{Delay Characterization}} of {{SCANeR Driving Simulator}}},
  author = {Saidi, François and Millet, Guillaume and Gallée, Gilles},
  date = {2010},
  langid = {english}
}

@article{sawadaEffectsSynchronisedEngine2020,
  title = {Effects of Synchronised Engine Sound and Vibration Presentation on Visually Induced Motion Sickness},
  author = {Sawada, Yuki and Itaguchi, Yoshihiro and Hayashi, Masami and Aigo, Kosuke and Miyagi, Takuya and Miki, Masayuki and Kimura, Tetsuya and Miyazaki, Makoto},
  date = {2020-05-12},
  journaltitle = {Scientific Reports},
  shortjournal = {Sci. Rep.},
  volume = {10},
  number = {1},
  pages = {7553},
  issn = {2045-2322},
  doi = {10.1038/s41598-020-64302-y},
  url = {https://www.nature.com/articles/s41598-020-64302-y},
  urldate = {2026-09-10},
  langid = {english}
}

@online{shahAirSimHighFidelityVisual2017,
  title = {{{AirSim}}: {{High-Fidelity Visual}} and {{Physical Simulation}} for {{Autonomous Vehicles}}},
  shorttitle = {{{AirSim}}},
  author = {Shah, Shital and Dey, Debadeepta and Lovett, Chris and Kapoor, Ashish},
  date = {2017},
  doi = {10.48550/ARXIV.1705.05065},
  url = {https://arxiv.org/abs/1705.05065},
  urldate = {2026-09-10},
  version = {2}
}

@article{silvaRealistic3DSimulators2024,
  title = {Realistic {{3D Simulators}} for {{Automotive}}: {{A Review}} of {{Main Applications}} and {{Features}}},
  shorttitle = {Realistic {{3D Simulators}} for {{Automotive}}},
  author = {Silva, Ivo and Silva, Hélder and Botelho, Fabricio and Pendão, Cristiano},
  date = {2024-09-10},
  journaltitle = {Sensors},
  volume = {24},
  number = {18},
  pages = {5880},
  issn = {1424-8220},
  doi = {10.3390/s24185880},
  url = {https://www.mdpi.com/1424-8220/24/18/5880},
  urldate = {2026-09-10},
  langid = {english}
}

@inproceedings{silveraDReyeVRDemocratizingVirtual2022,
  title = {{{DReyeVR}}: {{Democratizing}} Virtual Reality Driving Simulation for Behavioural \& Interaction Research},
  booktitle = {Proceedings of the 2022 {{ACM}}/{{IEEE}} International Conference on Human-Robot Interaction},
  author = {Silvera, Gustavo and Biswas, Abhijat and Admoni, Henny},
  date = {2022},
  series = {Hri '22},
  pages = {639--643},
  publisher = {IEEE Press},
  location = {Sapporo, Hokkaido, Japan},
  pagetotal = {5}
}

@inproceedings{simmannDesignAlternativeHardware2024,
  title = {Design of an {{Alternative Hardware Abstraction Layer}} for {{Embedded Systems}} with {{Time-Controlled Hardware Access}}},
  author = {Simmann, Gabriel and Veeranna, Vinay and Kriesten, Reiner},
  date = {2024-07-02},
  publisher = {SAE International},
  doi = {10.4271/2024-01-2989},
  eventtitle = {2024 {{Stuttgart International Symposium}}},
  langid = {english}
}

@online{skydrive_website,
  title = {Sky-{{Drive}}: {{A Distributed Multi-Agent Simulation Platform}} for {{Human-AI Collaborative}} and {{Socially-Aware Future Transportation}}},
  author = {{Sky-Drive Project}},
  date = {2025},
  url = {https://sky-lab-uw.github.io/Project%20SkyDrive/},
  urldate = {2026-09-10}
}

@misc{starkCooperativeDrivingAutomation2025,
  title = {Cooperative {{Driving Automation}} — {{Highway Driving Simulator Architecture}}: {{High-Level Architecture Design}}},
  shorttitle = {Cooperative {{Driving Automation}} — {{Highway Driving Simulator Architecture}}},
  author = {Stark, John and Lou, Yingyan and Yuan, Cheng and Slattery, Ethan and Nangle, Collin and Gayman, David and Chen, Eric and Rush, Kyle},
  date = {2025-05-01},
  url = {https://rosap.ntl.bts.gov},
  urldate = {2026-09-10},
  langid = {english}
}

@online{technologiesRDSModular2024,
  title = {{{RDS-Modular}}},
  author = {Technologies, Realtime},
  date = {2024},
  publisher = {Realtime Technologies},
  url = {https://www.faac.com/realtime-technologies/},
  urldate = {2026-09-10}
}

@inproceedings{vogetAUTOSARAutomotiveTool2010,
  title = {{{AUTOSAR}} and the Automotive Tool Chain},
  booktitle = {Proceedings of the {{Conference}} on {{Design}}, {{Automation}} and {{Test}} in {{Europe}}},
  author = {Voget, Stefan},
  date = {2010},
  series = {{{DATE}} '10},
  pages = {259--262},
  publisher = {{European Design and Automation Association}},
  location = {Leuven, BEL},
  isbn = {978-3-9810801-6-2},
  venue = {Dresden, Germany}
}

@inproceedings{wegenerTraCIInterfaceCoupling2008,
  title = {{{TraCI}}: An Interface for Coupling Road Traffic and Network Simulators},
  shorttitle = {{{TraCI}}},
  booktitle = {Proceedings of the 11th Communications and Networking Simulation Symposium},
  author = {Wegener, Axel and Piórkowski, Michał and Raya, Maxim and Hellbrück, Horst and Fischer, Stefan and Hubaux, Jean-Pierre},
  date = {2008-04-14},
  pages = {155--163},
  publisher = {ACM},
  location = {Ottawa Canada},
  doi = {10.1145/1400713.1400740},
  url = {https://dl.acm.org/doi/10.1145/1400713.1400740},
  urldate = {2026-09-10},
  eventtitle = {{{SCS SSM}}'08: {{Spring Simulation Multiconference}}},
  isbn = {978-1-56555-318-7},
  langid = {english}
}

@inproceedings{wuDontBlockGround2021,
  title = {Don’t {{Block}} the {{Ground}}: {{Reducing Discomfort}} in {{Virtual Reality}} with an {{Asymmetric Field-of-View Restrictor}}},
  shorttitle = {Don’t {{Block}} the {{Ground}}},
  booktitle = {Proceedings of the 2021 {{ACM Symposium}} on {{Spatial User Interaction}}},
  author = {Wu, Fei and Bailey, George S and Stoffregen, Thomas and Suma Rosenberg, Evan},
  date = {2021-11-09},
  series = {{{SUI}} '21},
  pages = {1--10},
  publisher = {Association for Computing Machinery},
  location = {New York, NY, USA},
  doi = {10.1145/3485279.3485284},
  url = {https://dl.acm.org/doi/10.1145/3485279.3485284},
  urldate = {2026-09-10},
  isbn = {978-1-4503-9091-0}
}

@article{yangUsingDistributedSimulations2024,
  title = {Using Distributed Simulations to Investigate Driver-Pedestrian Interactions and Kinematic Cues: {{Implications}} for Automated Vehicle Behaviour and Communication},
  shorttitle = {Using Distributed Simulations to Investigate Driver-Pedestrian Interactions and Kinematic Cues},
  author = {Yang, Yue and Lee, Yee Mun and Kalantari, Amir Hossein and De Pedro, Jorge Garcia and Horrobin, Anthony and Daly, Michael and Solernou, Albert and Holmes, Christopher and Markkula, Gustav and Merat, Natasha},
  date = {2024-11},
  journaltitle = {Transportation Research Part F: Traffic Psychology and Behaviour},
  shortjournal = {Transp. Res. Part F Traffic Psychol. Behav.},
  volume = {107},
  pages = {84--97},
  issn = {13698478},
  doi = {10.1016/j.trf.2024.08.027},
  url = {https://linkinghub.elsevier.com/retrieve/pii/S1369847824002341},
  urldate = {2026-09-10},
  langid = {english}
}

\newpage

\end{document}